\documentclass[italian, Lau, oneside, noexaminfo]{uaqthesis}

\usepackage[utf8]{inputenx}
\usepackage{tabularx}
\usepackage{indentfirst}
\usepackage{microtype}
\usepackage{amsfonts}
\usepackage{algorithm}
\usepackage{algorithmic}
\usepackage{verbatim}

\usepackage{lettrine}
\usepackage[nottoc, notlof, notlot]{tocbibind}
\usepackage{hyperref}

\hypersetup{
			colorlinks=true,
			linkcolor=black,
                        linktoc=page,
			anchorcolor=black,
			citecolor=black,
			urlcolor=black,
			pdftitle={My Beautiful Thesis},
			pdfauthor={Name Surname},
			pdfkeywords={thesis, univaq, l'aquila, university}
 }

\usepackage{indentfirst}
\usepackage{todonotes}
\usepackage{graphicx}
\usepackage{float}
\usepackage{csquotes}
\usepackage{amsmath}
\usepackage{numprint}
\usepackage{pgfplots}
\usepackage{pgfplotstable}

\usepackage[toc]{appendix}

\usepackage{inconsolata}
\usepackage{color}
\usepackage{listings}

\graphicspath{{./images/}}
\pgfplotsset{compat=1.14}

\definecolor{pgreen}{rgb}{0,0.5,0} 
\definecolor{pred}{rgb}{0.9,0,0} 
\definecolor{javared}{rgb}{0.6,0,0} 
\definecolor{javagreen}{rgb}{0.25,0.5,0.35} 
\definecolor{javapurple}{rgb}{0.5,0,0.35} 
\definecolor{javadocblue}{rgb}{0.25,0.35,0.75} 

\definecolor{forestgreen}{rgb}{0.0, 0.27, 0.13}

\pgfplotstableset{
    every head row/.style={before row=\toprule,after row=\midrule},
    every last row/.style={after row=\bottomrule},
    fixed,precision=2,
}

\title{Control of Automated Driving in Motion Planning}
\author{Yucheng LI}
\IDnumber{274556}
\course[]{Computer and Systems Engineering}
\courseorganizer{Department of Computer and Systems Engineering}
\submitdate{2022}
\copyyear{2022}
\advisor{Prof. Stefano Di Gennaro, }
\coadvisor{Prof. Maarouf Saad}
\authoremail{yucheng.li@student.univaq.it}

\begin{document}

\frontmatter
\maketitle
\dedication{}



\tableofcontents

\mainmatter

\renewcommand\thesection{\arabic {section}}

\section*{Abstract}
Autonomous driving vehicles aim to free the hands of vehicle operators, helping them to drive easier and faster, meanwhile, improving the safety of driving on the highway or in complex scenarios. Automated driving systems (ADS) are developed and designed in the last several decades to realize fully autonomous driving vehicles (L4 or L5 level).

Usually, ADS can be divided into three main processes: Perception, Motion Planning and Control. Although many prototypes of ADSs were already implemented, both in scientific research centres and industries, a lot of work has still to be done, due to the complexity of different driving scenarios and the need for real-time computation. In particular, motion planning still remains one of the hardest problems to solve, since it deals with huge data from perception, and with the problem of finding (in the fastest possible manner) an optimal trajectory, to be used as a reference for the controller. Therefore, the high computational request is demanded of the CPUs onboard the autonomous vehicles, and planning algorithms with high performance have to be designed, to realize the real-time computation. Failing to find a solution in real-time could lead to a non-optimality of the reference trajectory, and to potential traffic accidents.

The scale of sampling space leads to the main computational complexity. Therefore, by adjusting the sampling method, the difficulty to solve the real-time motion planning problem could be incrementally reduced. Usually, the Average Sampling Method is taken in Lattice Planner, and Random Sampling Method is chosen for RRT algorithms. However, both of them don't take into consideration the prior information, and focus the sampling space on areas where the optimal trajectory is previously obtained. Therefore, \emph{in this thesis it is proposed an adaptive sampling method to reduce the computation complexity, and achieve faster solutions while keeping the quality of optimal solution unchanged}. 

The main contribution of this thesis is the significant decrease in the complexity of the optimization problem for motion planning, without sacrificing the quality of the final trajectory output, with the implementation of an Adaptive Sampling method based on Artificial Potential Field (ASAPF). In addition, also the quality and the stability of the trajectory is improved due to the appropriate sampling of the appropriate region to be analyzed.

To prove the feasibility of the proposed ASAPF, the whole processes of Automated Driving including Perception, Motion Planning and Control have been simulated in MATLAB/Simulink, used in combination with CARSim and Prescan, which are popular simulation and visualization software for Automated Driving. The simulation results prove that the proposed ASAPF is feasible and efficient, compared with conventional Average Sampling Methods for the discretization of the continuous space, used as input to the dynamic programming analysis. Comparing the proposed ASAPF with classical Average Sampling Methods, the computational time necessary to solve the same optimization problem in real-time reduces by around 2/3, which could significantly improve the efficiency of the motion planning algorithm, and the safety of automated driving.

\section*{Keywords}
\noindent Automated Driving, Motion Planning, Optimization Function, Adaptive Sampling, Artificial Potential Field, Dynamic Programming, Quadratic Programming, Nonlinear Control System,  Double PID control, LQR control,  MATLAB/Simulink, CARSim.

\newpage
\section{Introduction}
Autonomous driving has promised to improve drastically the convenience of driving by releasing the burden of drivers and reducing traffic accidents with more precise control. According to a past technical report by the National Highway Traffic Safety Administration(NHTSA), 94 per cent of traffic accidents are mainly caused by human manipulating errors \cite{singh2015critical}. Nowadays, with the fast development of artificial intelligence and significant advancements in the Internet of Things (IoT) technologies, we have witnessed the steady progress of autonomous driving over the recent years. In \cite{montgomery2018america}, it is mentioned that annual social benefits of ADSs are projected to reach almost \$800 billion by 2050 through road casualty reduction, decreased energy consumption and increased productivity caused by the reallocation of driving time.

The earliest major automated driving study is Eureka Project PROMETHEUS which was carried out in Europe from 1987-1995. In this project, VITA II by Daimler-Benz succeeded in automatically driving on highways \cite{ulmer1994vita}. DARPA Grand Challenge, organized by the US Department of Defense in 2004, was the first major automated driving competition. However, at this time, all the attendees failed to finish the 150-mile off-road parkour because it is difficult for the cars to drive automatically without any human intervention at any level. Urban scenes were seen as the biggest challenge of the field for ADSs, therefore, DARPA Urban Challenge \cite{buehler2009darpa} held in 2017 is significant for this field. In this competition, many different groups from all over the world implemented their ADSs in a test urban environment. Finally, six teams completed the event successfully. Even if this competition was the biggest and most significant event up to that time, the test environment was still not real because of the lack of certain aspects of a real-world urban driving scene such as pedestrians and cyclists. After DARPA Urban Challenge, several more automated driving competitions such as \cite{broggi2012vislab}-\cite{englund2016grand} were held in different countries.

\begin{figure}[!t]\centering
	\includegraphics[width=13.5cm]{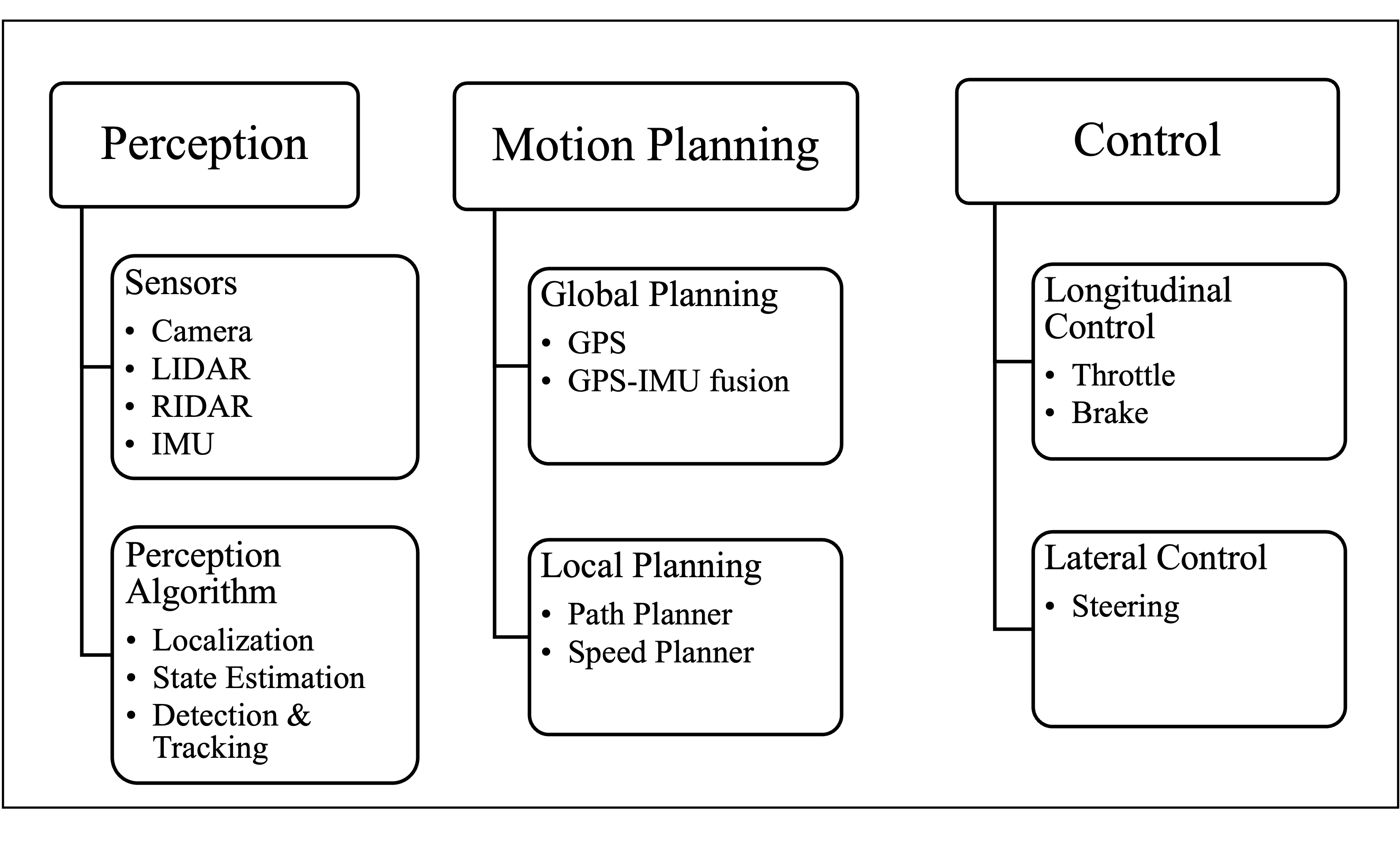}
	\caption{System Architecture of an Autonomous Vehicle}\label{sub systems}
	\end{figure}

Automated driving is the latest trend in the automobile industry with the enhancement of computational ability and improvement of sensor accuracy. It has been one of the most popular topics in academic institutes and high technology companies all over the world, especially in the United States of America, China and the United Kingdom. The concept of vehicle automation was envisioned in 1918 \cite{pendleton2017perception}.  In \cite{samak2021control}, the system structure of automated driving is explicitly analysed and autonomous vehicles realize the process of autonomous driving by three steps: perception, planning and control, as shown in Fig \ref{sub systems}. 

The main focus of our work is related to the planning and control components of the architecture, specifically local planning techniques in automated vehicles, for the reason that it is one of the most difficult problems to be solved in automated driving techniques due to the unpredictable manoeuvres of other dynamic vehicles, pedestrians and other traffic participants. A review of the different approaches and concept definitions (as global, local or reactive motion planning) can be found in \cite{kunchev2006path}, \cite{hwang1992gross}, \cite{elbanhawi2014sampling}. Normally, these planning techniques can be classified into four main groups depending on their techniques and implementation in automated driving: Graph Search Based Planners, Sampling Based Planners, Interpolating Curve Planners and Numerical Optimization. In the beginning, as for the automated driving problem, the basic idea is to transfer it into a graph problem. To be specific, the state space is traversed into a graph or an occupancy grid. Then, graph-searching algorithms can be implemented so as to find a possible solution. Some of these algorithms, for example, Dijkstra Algorithm in \cite{marchese2006multiple}, \cite{lavalle1998optimal}, A-Star Algorithm (A*) \cite{likhachev2009planning}, Dynamic A* Algorithm \cite{stentz1997optimal} and State Lattice Algorithm \cite{pivtoraiko2005efficient} have been successfully applied to the automated vehicles development. The easiest and most common graph-searching algorithms are Depth-First Search (DFS) and Breadth-First Search(BFS). BFS uses stacks to save the visited nodes, therefore it needs less execution time compared with DFS which doesn't use stacks, but scarifying more storage space. Dijkstra is another popular and simple graph-searching algorithm, which tries to find the single-source shortest path in the graph. The configuration space is approximated as a discrete cell-grid space, lattices. A-Star Algorithm is a type of improvement of the Dijkstra Algorithm that can realize a very fast node-searching due to the design of heuristics. The most important design aspect is the determination of the cost function, which will directly decide the quality of performance of path planning. Rapidly-Exploring Random Tree (RRT) is the most commonly used method in robotics for motion planning \cite{lavalle2001randomized}, so many researchers also tested this algorithm for automated vehicles. For example, the MIT team used it for automated driving at DARPA Urban Challenge in \cite{kuwata2009real}. However, the fatal drawback of this algorithm is that the resulting path is not optimal, jerky, and even not curvature continuous. The passengers will feel uncomfortable if we implement this algorithm for motion planning. RRT* in \cite{kuwata2009real} got improved based on basic RRT and developed by the same team in MIT to make sure it can converge to an optimal solution. Nevertheless, the drawback we mentioned above in traditional RRT still exists in this improved RRT* algorithm. Interpolation is defined as the process of constructing and inserting a new set of data within the range of a previously known set, so it is mainly used to do smoothing tasks with a given reference line so as to satisfy vehicle constraints, and trajectory continuity and improve the performance of planning trajectory. In conclusion, the classification and their main representative algorithms are listed in the table\ref{comparison_table}. In addition, the pros and cons of them are also given in the same table so as to have a better comparison.

\begin{table*}[!htbp]
    \centering
    \scriptsize
    \begin{tabularx}{\textwidth}{lXXX}
    \toprule
      Methodologies   & Examples & Advantage & Disadvantage \\
      \toprule
      Graph Search Based   & Dijkstra, A*, D*, State Lattices & Transform to a graph problem and efficiently use classic graph search algorithms
      & Slow due to vast areas and works only with a full-known map which is impossible in real-time implementation
      \\ 
      \midrule
      Sampling-Based & RRT, RRT*, RRT-connect & Provide a fast solution in multi-dimensional systems. Complete algorithm and always converges to a solution (if existed)
      & The resulting trajectory is not continuous and therefore jerky
      \\
      \midrule
      Interpolating Curves & Polynomial Curves, Bezier Curves,
      Spline Curves & Optimize the curvature and smoothness of the path with CAGD techniques
      & Depends on global planning or waypoints. Time-consuming to deal with obstacles in real time
      \\
      \midrule
      Numerical Optimization & Function Optimization & Road and ego-vehicle constraints as well as other road users can be easily taken into account
      & Time-consuming, therefore, the optimization should be stopped at a given time horizon. 
      \\
      \bottomrule
    \end{tabularx}
    \caption{Comparison of four different mainstream motion planning methodologies}
    \label{comparison_table}
\end{table*}
                                 
From the table \ref{comparison_table}, we could see that all these algorithms for motion planning have their advantages and disadvantages. That means, until now, there is no one algorithm outperforming others, so we need to choose one from them or combine them together to implement our motion planning task depending on the demands of our real implementation problem. After considering the pros and cons of different motion planning technologies and their ability to implement in the virtual or real-time environment, Baidu Apollo EM Planner decided to take an optimization function combined with spline curves to generate a trajectory, \cite{fan2018baidu}, which could satisfy the curvature request that is important for the control module to do path following and meet the mission of low computational task, that means there is not too much demand for the high ability CUP onboard autonomous vehicles. 
On the one hand, spline curves, specifically polynomial curves need low computational cost compared with graph searching algorithms, especially the RRT family because of their random sampling. On the other hand, the optimization function can make sure the trajectory is smoothing enough to satisfy the request of comfort, vehicle dynamic constraints and so on. 

Specifically speaking, firstly continuous construction space needs some methods of discretization, for example, random sampling, average sampling, etc. In this thesis, to solve this problem and improve the efficiency of the motion planning algorithm, we proposed our Adaptive Sampling Method based on Artificial Potential Field (ASAPF). Then based on the discrete space, Dynamic Programming (DP) is used to obtain a rough solution. After achieving the rough solution, a convex space is built around it. Then Quadratic Programming (QP) is chosen to explore an optimal solution inside the convex workspace. The solution will converge to local optima if we iteratively implement the process of path planning and speed planning. This whole motion planning part can be finished within 100 ms, therefore, it can satisfy the real-time requirement. However, with our proposed ASAPF, the complexity of the optimization problem will be reduced by adjusting the density of sampling with the help of prior information (like optimal trajectory in the last cycle), obstacles information and road boundaries and some unnecessary sampled points will be deleted before QP. Therefore, the computational time will be further shortened, meanwhile, the efficiency of our motion planning algorithm in automated driving and the safety of autonomous vehicles will be incrementally improved.

The structure of this thesis is shown below: Section \ref{Baidu Apollo and our whole methodology} is the basement of our motion planning. We take Baidu Apollo EM Planner as our reference methodology, analysing the advantages and drawbacks of their algorithm. Then, our methodology (ASAPF) is proposed to solve their problem and reduce the computational complexity so as to improve the real-time implementation ability of motion planning. Section \ref{Motion Planning} introduces the details of our ASAPF and the formulation of DP and QP in the motion planning algorithm. Section \ref{Vehicle Models} gives the formulation of simplified dynamic vehicle models (bicycle model) which are commonly used for automated driving and important for the analysis of control algorithm. Section \ref{Control Algorithm} shows the control algorithm we implemented in the control module of the automated driving system. Linear Quadratic Regulator (LQR) and PID are chosen for the lateral and longitudinal control separately because of their stability and simplicity in the control area. Section \ref{Simulation and results} firstly explains our simulation environment (Matlab/Simulink + CARSim + Prescan) and lists the hyperparameters we used in our Simulink model. Then, in this section, the results of the simulation are shown and the comparison of the running time of DP with the original Average Sampling and our proposed ASAPF is analyzed so as to prove the efficiency and feasibility of our method. Section \ref{Conclusion and future works} concludes what we did in this master thesis and analyses some aspects that we didn't consider here due to the limitation of time. Then I propose some possible directions on which I can choose to research in my PhD. period in the future.

\section{Baidu Apollo and our whole methodology}
\label{Baidu Apollo and our whole methodology}
Baidu which starts as a search engine company in China (like Google Searching Engine) has already been a leading AI company with a strong Internet foundation now. Since 2013, they have already been working on automated driving. After four years of researching this area, they officially published their automated driving open-source platform named Apollo in  \href{https://github.com/ApolloAuto/apollo}{\textsl{Baidu Apollo Github}}. Apollo is an open-source Automated Driving Platform, so everyone who is interested in automated driving could use it to study automated driving technologies and even implement it in Linux to visualize it in a virtual environment or in real vehicles. As for the motion planning module, after doing research on the existing methodologies and technologies, they decided to use the Optimization function combined Polynomial Curve. In other words, Dynamic Programming and Quadratic Programming are realized in their motion planning separately for path planning and speed planning. Then the path profile and speed profile are merged together to generate the optimal trajectory avoiding static and dynamic obstacles at the same time in the real traffic road. Their motion planning method is called EM Planner, because Expectation and Maximization are iterative used inside, which is published out in \cite{fan2018baidu}. Compared with graph searching methodologies, the trajectory is less jerky and easier to realize the real-time path planning. In this chapter, the main structure of Apollo EM Motion Planner will be introduced and some problems existing in the real-time implementation will be mentioned. In the end, we will propose our method (ASAPF) and how it could solve the existing problems. 
\subsection{Frenet Coordinate}
Nowadays, most autonomous driving motion planning tasks are developed in Frenet frames with time (SLT) so as to reduce the planning dimension with the help of a reference line, like in \cite{werling2010optimal}, \cite{gu2012road} and \cite{peng2022motion}. The reason that Frenet frames are very popular in Automated Driving is the expression of equations will become simpler in Frenet Coordinate System and it will be easier to decouple the lateral and longitudinal motion planning and control.
\begin{figure}[!t]\centering
	\includegraphics[width=13.5cm]{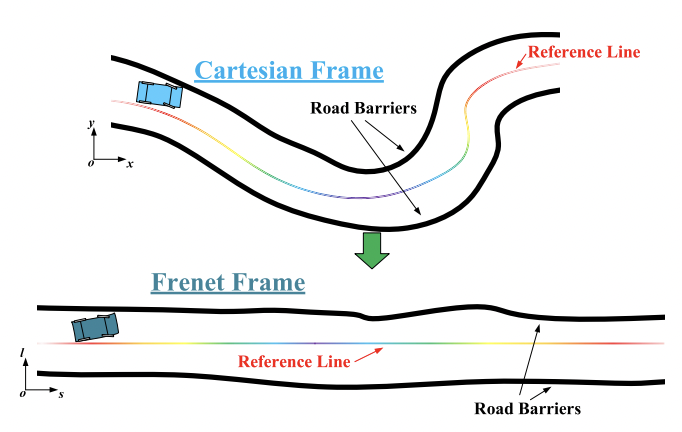}
	\caption{ Schematics on the conversion from Cartesian frame to Frenet frame}
	\label{Cartesian_Frenet}
	\end{figure}
As depicted in Figure \ref{Cartesian_Frenet}, the same road in the real world looks totally different in the Cartesian frame and in Frenet Coordinate System \cite{li2022autonomous}. It seems like a straight road in the latter frame after converting the road with an irregular shape from a Cartesian to a Frenet frame. It is obvious that any road could be standardized as a straight tunnel using a Frenet frame. Then, the nonlinear collision-avoidance constraints will be transformed into linear within-tunnel constraints. In addition, the originally coupled kinematic constraints are decoupled as independent polynomials in the longitudinal and lateral dimensions \cite{li2015real}. Then, we could easily solve our Dynamic Programming separately in the SL graph and ST graph within the Frenet frame, which seems impossible in the Cartesian coordinate system. To be specific, it is still difficult to solve a 3-Dimension SLT problem, therefore first of all, the SLT problem can be easily decoupled into two 2-Dimension problems (SL and ST). SL is the relation between the station and lateral distance. ST represents station and time. By solving the SL problem, we could get the path profile which can make sure the static and low-speed obstacles avoidance. Then, the next step is to find the relation between station and time which will create the speed profile and it can offer us a solution to avoid dynamic obstacles. Because both of path profile and speed profile have the same dimension S (station), they can be merged together by matching the station coordinate system introduced in \cite{fan2018baidu} and until now, we obtain the final trajectory which can be used as input for control module to do the path following. 
\subsection{EM planner Iteration}
\begin{figure}[!t]\centering
	\includegraphics[width=13.5cm]{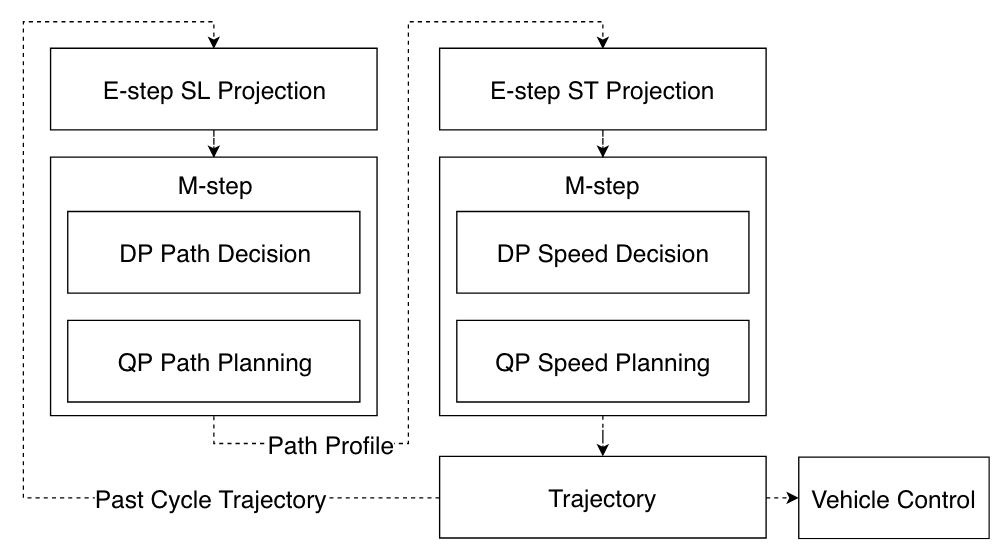}
	\caption{EM Planner Iteration \cite{fan2018baidu}}\label{EM Planner}
	\end{figure}
As we introduced above, the 3-Dimension problem (SLT) should be divided into two 2-Dimension problems (SL and ST), and then these two problems can be optimized separately and iteratively, as shown in \ref{EM Planner}. Let us consider the SL graph first. In E-step, observed obstacles are projected in the SL graph and we need to find the optimal solution considering obstacles, dynamic vehicle constraints and traffic rules. The easiest way to solve this problem is quadratic programming which is similar to Newton's gradient descent. However, as we all know, quadratic programming can give us a fast and good solution only in convex space. However, obviously, motion planning is not a convex problem. Thus, in the Apollo EM planner, dynamic programming is used firstly to obtain a rough solution and this solution will provide decisions or manoeuvres like nudge, yield or overtake. Then the next step is to build a convex hull based on that rough decision. Until now, convex space is already obtained and it can be used for quadratic programming to find a fast and optimal solution. Secondly, the ST graph is built to consider dynamic obstacles. The dynamic obstacles are mapped with the help of the last cycle trajectory of the ego car. The main idea is to project the last cycle's moving trajectory on the Frenet frame so as to extract the station direction speed profile. That means we could more easily obtain an estimate of the ego car's station coordinates given a specific time, which will be used to evaluate the interactions with dynamic obstacles. Once an ego car's station coordinates have interacted with an obstacle trajectory point at the same time, a shaded area on the ST map will be marked as the estimated iteration with a dynamic obstacle. ST projection helps us generate the ego car's speed profile so as to realize dynamic obstacle avoidance. Then, the final optimal trajectory is generated by the combination of the path profile and speed profile.
\subsection{Existed problems and our improvement}
As analyzed above, Quadratic Programming is a good method to obtain a fast and optimal solution. However, it only works on the convex space. Therefore the non-convex space needs to be transformed into convex space. In EM Planner, Dynamic Programming is implemented to search for a rough solution first and build a convex hull around it. In theory, this method seems feasible. However, in real implementation, due to the complexity of computation of Dynamic Programming, this method hugely increases the running time of programming and sometimes it is hard to obtain the solution in real-time which will lead to the failure of motion planning. Autonomous vehicles always work at high speed with other dynamic vehicles, pedestrians and bicycles in complex scenarios like urban and highways where the failure of real-time motion planning will lead to serious traffic accidents, even threatening the life of human beings. After analyzing the problem and the methodologies, we find the problem which makes the computation of Dynamic Programming heavy is Average Sampling and the computational cost of some points that the optimal trajectory will never pass through. To solve this problem, what we consider is whether we could delete some unnecessary points before Dynamic Programming optimization, so as to simplify the computational task. How to distinguish these unnecessary points from useful points becomes a difficult point that we need to consider, because we can not scarify the performance of our final trajectory to save the running time. Then, we have the idea of taking advantage of prior information (prior optimal trajectory), because the optimal trajectory in the next cycle and that in the last cycle should not change a lot because of the constraints of dynamic vehicles and the request for the comfort of passengers. Another reason is the time interval between two motion planning is only 0.1 second which is too small to have big vibration of trajectory in the normal driving condition. Therefore, the optimal trajectory in the next cycle should be close to the trajectory in the last cycle. In addition, the optimal trajectory should not be too close to road boundaries and obstacles to make sure the requested safety. Inspired by Artificial Potential Field (APF), the obstacles and road boundaries are considered as repulsive forces and the optimal path in the last cycle can create an attractive force field. Then, we could rank the points in the same station by their value given by repulsive and attractive fields. The next step is to choose the minimal several points (for example, 5 points in our experiment) to create an adaptive sampling space for dynamic programming to find the rough solution. If the row and column of Average Sampling in motion planning are 11 and 4 for example, then the complexity of computation is $\mathcal{O}(4*11^2)$. After adopting our ASAPF, the complexity becomes $\mathcal{O}(4*5^2)$ while keeping the performance with the same quality. Our experiment in Simulink shows the running time of Dynamic Programming in ASAPF is normally reduced by 2/3 in comparison with the original Average Sampling Method and proves the feasibility of our proposed method. With the help of our adaptive sampling method, the computational request is largely reduced, so it will be faster to obtain the solution and the request for a high-performance CPU is decreased because the computational task is simplified before dynamic programming. The most important thing is, the performance of the final trajectory is not sacrificed and will even get improved if we increase the density of longitudinal sampling with our proposed ASAPF.  

\newpage

\section{Motion Planning}\label{Motion Planning}

Motion planning can be divided into two components: path planning and speed planning (i.e., speed profile generation). Then, the path and speed profile are merged together to generate the final trajectory. There are many conventional sampling-based algorithms for path planning in motion planning that is proven feasible and useful, for example, Rapidly-Random exploring Tree (RRT) in \cite{4651075} and RRT* in \cite{noreen2016optimal} (an improved method by the same team), Probabilistic Roadmap (PRM) in \cite{song2003general}, etc. Although they are in general computationally efficient to find a feasible path, these random sampling-based methods tend to fail to find the optimal path in most of the scenarios mentioned in \cite{li2020adaptive}. Nowadays, the popular methods to implement motion planning are based on the Frenet coordinate system, which means we need to transform the traditional Cartesian coordinate system to the Frenet coordinate system first, and then do the motion planning in the Frenet coordinate system. After receiving the final trajectory from the motion planning part, the last step is to transform the trajectory back from Frenet to Cartesian so as to follow the path easier for the control part. According to Lindemann and LaValle (2005) \cite{lindemann2005current}, motion planning is an NP-hard problem following the standard of computer and algorithm, but randomization could make it easier to solve this type of finding a feasible path in even non-convex space. Due to the limitation of computational ability in the onboard CPU, there should be a trade-off between the time efficiency of the algorithm and the optimality of the solutions. Definitely, higher sampling resolution will help to find better solutions to motion planning problems, however, it will also increase the computational complexity to solve the problem at the same time, leading to dangerous manoeuvres for autonomous vehicles in real-time implementation. In \cite{ichter2018learning}, Ichter et al. (2018) proposed a method which is to uses a non-uniform resolution to replace the uniform resolution sampling to solve this problem in RRT-based motion planning.  In \cite{kumar2012adaptive}, Kumar and Chakravorty (2012) introduced an adaptive-resolution-based sampling strategy in Probabilistic Roadmap (PRM) by explicitly utilizing the information encoded in the connectivity graph. More recently, in 2020 IFAC \cite{li2020adaptive}, Zhaoting Li from HIT, China implemented an adaptive sampling method to determine the trajectory by dynamic programming. However, they only used reference line and obstacles information to adjust the density of sampling, disregarding the information from history information, therefore, the performance and the running time didn't get too much improved. In their future work, they also suggested that history information should be used to dynamically adjust sampling density \cite{li2020adaptive}. Based on their drawback and experience, we proposed our adaptive sampling method based on historical information to save the running time of dynamic programming in real-time.

\begin{figure}[!t]\centering
	\includegraphics[width=13.5cm]{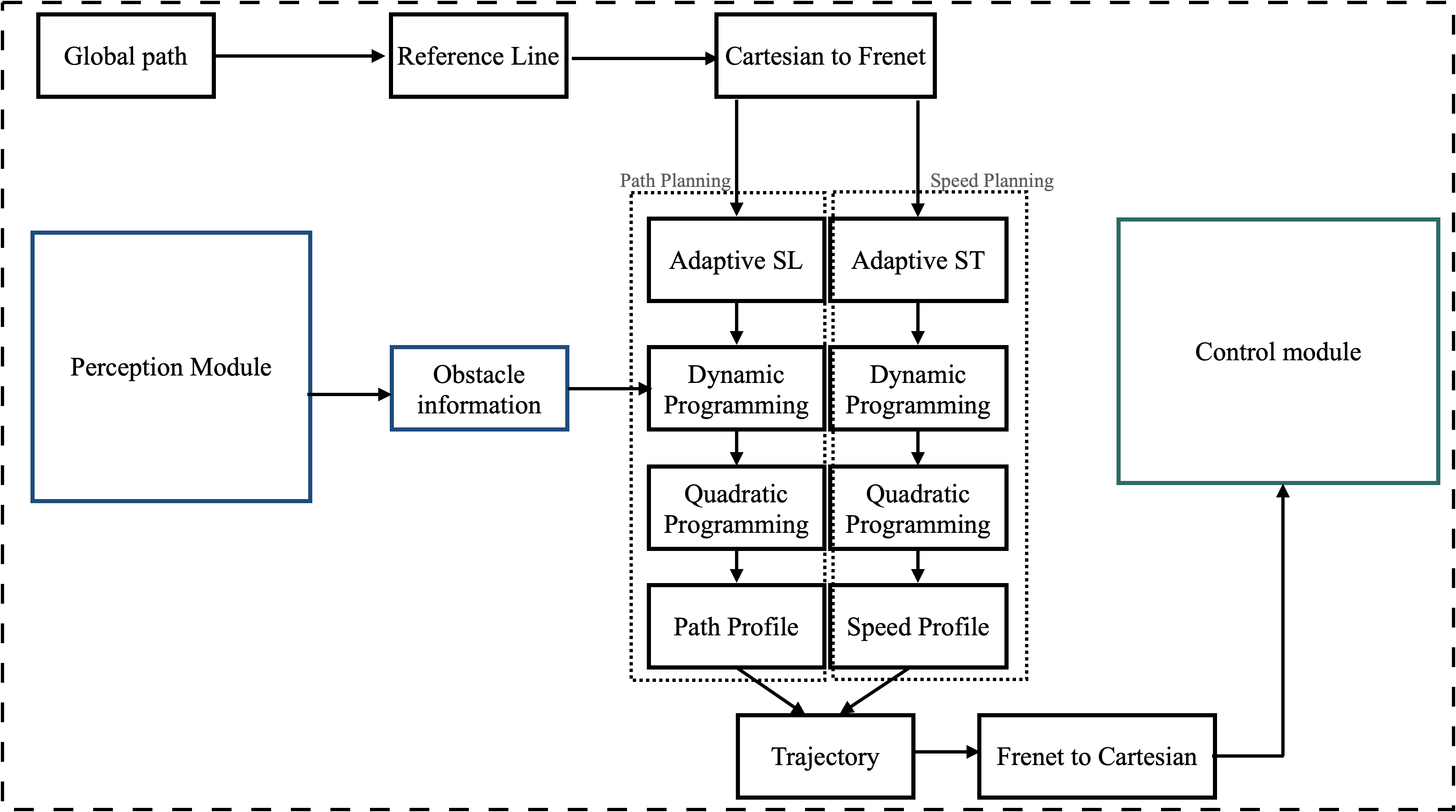}
	\caption{Overview of our motion planning framework }\label{framework}
	\end{figure}
	
Here is the structure of my method in Figure \ref{framework}. Initially, we admitted that the global path has already been provided by the upper layer, so we could focus on the local planning part to avoid obstacles. The global path only roughly tells us which way we could choose to go, like the information offered by GPS navigation, but not how to go exactly and how to avoid obstacles and follow traffic regulations. Therefore, we need the local planner to achieve the final trajectory. The global path is always too long and too jerky to guild autonomous vehicles to drive in real time. That's why the first step is to 
cut the global path into the local path and smooth it into a reference line, which will be discussed more specifically later. The reference line is represented in a Cartesian coordinate system, which is not easy to decouple and solve our motion planning mission. To solve this problem, the common method is to transfer the reference line from the Cartesian to the Frenet coordinate system. Then, the 3-Dimension problem (SLT) in the Frenet frame could be easily decoupled into two 2-Dimension problems (SL and ST) so as to simplify the problem because it is much easier to solve two 2-Dimension problems than solving a 3-Dimension optimization problem. In the next, adaptive sampling methods based on historical experience are implemented both in the SL graph and ST graph, and the computational task is incrementally simplified without scarifying the performance of the planning trajectory. Path planning modules including Dynamic Programming (DP) and Quadratic Programming (QP) will build path profiles, which will be used as the initialization for the ST graph, and then speed planning will offer us a speed profile after DP and QP. The last step is to merge path and speed profiles to generate the final planned trajectory and transform it back from Frenet to the Cartesian coordinate system because it is easier for the control module to do path following in the Cartesian coordinate system.

\subsection{Reference line}
\begin{figure}[!t]\centering
	\includegraphics[width=13.5cm]{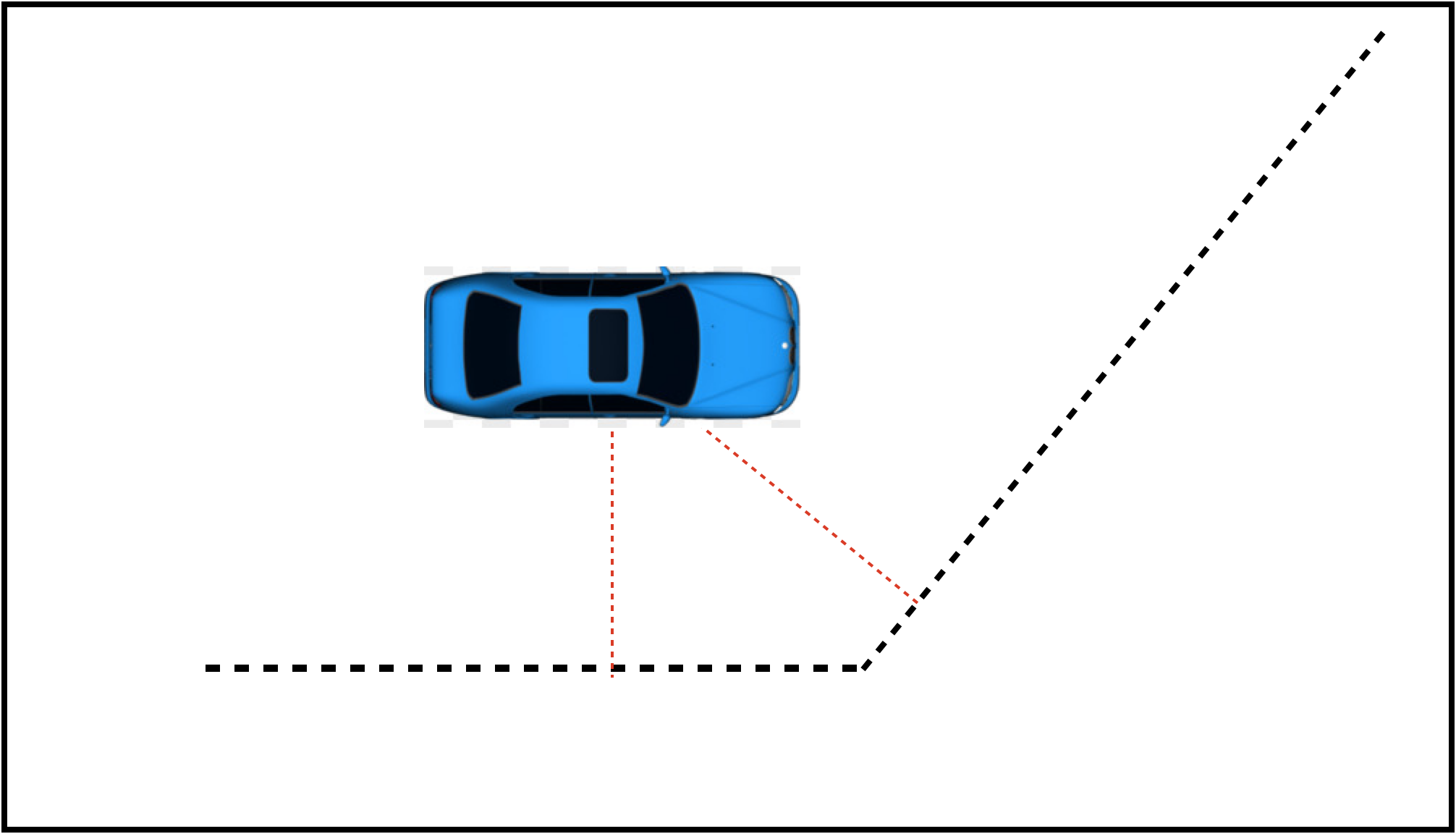}
	\caption{Not unique projection in global route}\label{global route}
	\end{figure}

The reference line is the link between global planning and local planning. Global planning offers guidance to the latter and local planning specifies how to avoid collision with obstacles on the road in real-time. There are mainly three reasons why a reference line is necessary and the global route is impossible to be directly used for local planning. The first reason is that the global route is too long to follow (for example, our GPS could offer us a way around 100 km from L'Aquila to Rome) so the computational task will become too heavy for the CPU with the global route. The easiest way to solve this problem is to prune the global route and generate a reference line. Secondly, our ego vehicle and obstacles need to be projected in the same line together so as to compare their positions and avoid collision between them. However, if we consider the global route as our reference line, the projection of ego vehicle and obstacles may not be unique which will make the failure of collision avoidance, like the example in Figure \ref{global route}. The last reason is that the global route is too jerky (smoothing enough), so the passengers will not feel comfortable in autonomous vehicles with that kind of jerky trajectory. The generated reference line will offer us a smoother line as input for local planners. The normal way to create a smoothing reference line is Quadratic Programming (QP).
The cost function (QP) includes three parts: reference cost, smooth cost and length cost, so we have the cost function equation:
\begin{equation}
   C_{reference\_line} = C_{ref} + C_{smooth} + C_{length} 
\end{equation}

\begin{figure}[!t]\centering
	\includegraphics[width=13.5cm]{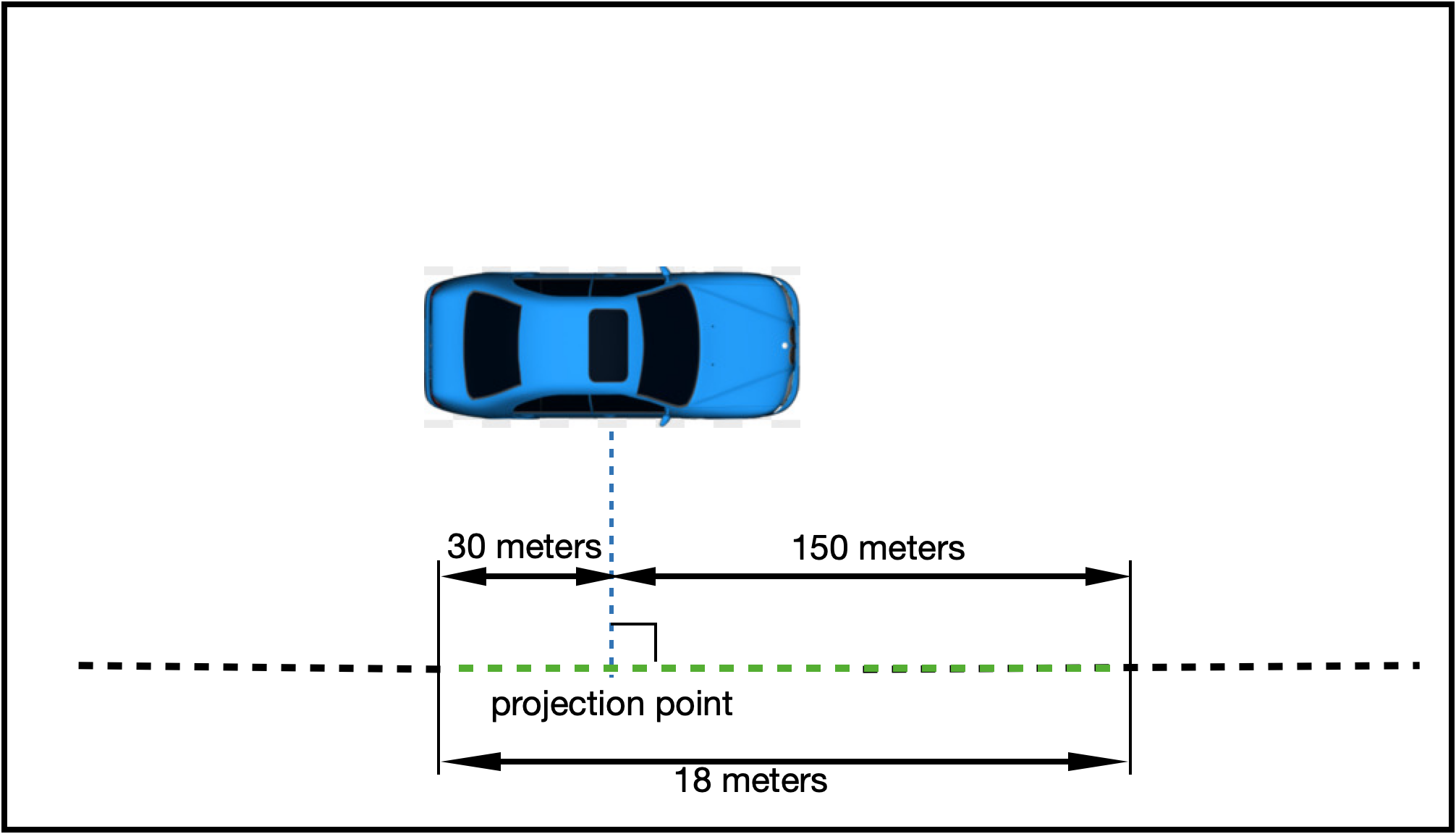}
	\caption{Projection and length of reference line}\label{reference line}
	\end{figure}

This cost function above is used only for three neighbour points starting from the first point in the global route every time. It will be iteratively implemented until the length of the reference line is enough. Normally, in our experiment, the length we choose to use is 180 meters. The length before the projection point of the ego vehicle is 150 meters, and the length after is 30 meters, as shown in Figure \ref{reference line}.
	
The reference line should be smoother than the original global route, but the deviation between them should not be too much, therefore reference cost is important here and the formulation is given below:
\begin{equation}
    C_{ref} = Coeff_{ref}*\sum_{i=1}^3((x_i - x_{ir})^2 + (y_i - y_{ir})^2)
\end{equation}
Here $x_{ir}$ and $y_{ir}$ are the  coordinates of discrete points on the reference line.

The smoothing function means the second point should try to keep not too far from the middle of the first and third points, therefore the equation of smoothing cost is given below:
\begin{equation}
    C_{smooth} = Coeff_{smooth}*((x_1 + x_3 - 2x_2)^2 + (y_1 + y_3 - 2y_2)^2)
\end{equation}

The last cost is length cost, which means the points in the reference line should be as close as possible to neighbour points:
\begin{equation}
    C_{length} = Coeff_{length}*\sum_{i=1}^3((x_{i+1} - x_i)^2 + (y_{i+1} - y_i)^2)
\end{equation}

\subsection{Cartesian and Frenet frame Transformation}
\begin{figure}[!t]\centering
	\includegraphics[width=13.5cm]{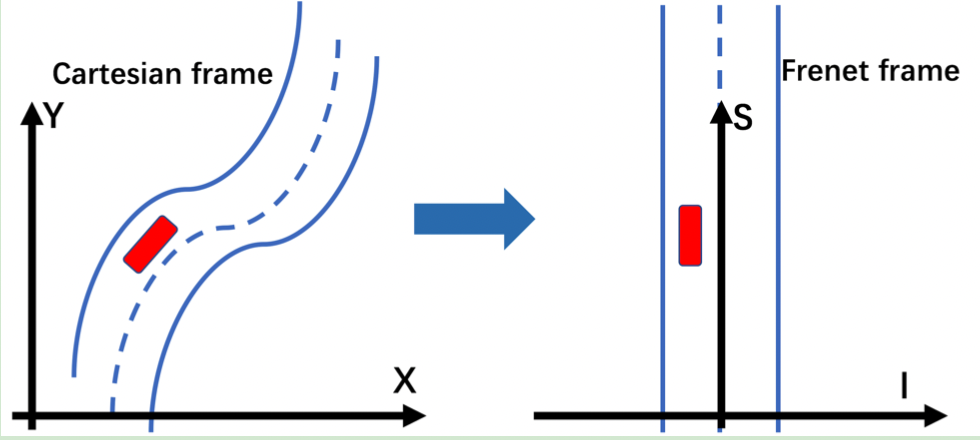}
	\caption{Cartesian to Frenet Transformation}\label{Cartesian to Frenet}
	\end{figure}
	
In this section, the Coordinate system transformation problem is mainly discussed. As shown in Figure \ref{Cartesian to Frenet}, the curve road boundaries in the Cartesian frame will be transformed into straight in the Frenet frame and the nonlinear constraints will become linear which is easier to solve our optimization function in motion planning task. In addition, the longitudinal distance will be presented as S (Station) and the lateral distance with reference to S will be shown as L (Lateral), so the problem will be decoupled into two separate directions. Assume we have $\Vec{r_n}$, $\Vec{v_n}$, $\Vec{a_n}$, $k_n$ in the Cartesian frame, the objective of transformation is to obtain the coordinates based on the reference line in Frenet coordinate system. Specifically speaking, we need to obtain $s$, $\dot{s}$, $\ddot{s}$, $l$, $l^{'}$,$l^{''}$, $\dot{l}$, $\ddot{l}$. The definition of these variables is given below:

\begin{equation}
    \dot{s} = \frac{ds}{dt},
    \ddot{s} = \frac{d^2s}{dt^2},
    l^{'} = \frac{dl}{ds},
    l^{''} = \frac{d^2l}{ds^2},
    \dot{l} = \frac{dl}{dt}, 
    \ddot{l} = \frac{d^2l}{ds^2}
\end{equation}

The process of how to transform is a little complex, so we put the proof of Cartesian to Frenet frame using a vector method in Appendix \ref{Appendix 2} and the Frenet to Cartesian is ignored here. Here the result is directly given in this section.
\begin{itemize}
    \item Cartesian to Frenet \\
    \begin{equation}
\left\{\begin{array}{lc}
s= & s_{r} \\
\dot{s} = & \frac{\dot{v}\vec{\tau_r}}{1-k_rl} \\
\ddot{s}= & \frac{\vec{a} \cdot \overrightarrow{\vec{\tau}_r}}{1-k_{r} \cdot l}+\frac{k_{r} \cdot \dot{s}^{2} \cdot l^{\prime}}{1-k_{r} l}+\frac{\dot{s}^{2}}{1-k_{r} l}\left(k_{r}^{\prime} \cdot l+k_{r} \cdot l^{\prime}\right) \\
l = & (\Vec{r_n} - \Vec{r_r})\Vec{n_r} \\
\dot{l} = & \vec{v} \cdot \vec{n_r}  \\
l^{'} = & (1-k_rl)\frac{\vec{v}\vec{n_r}}{\vec{v}\vec{\tau_r}}  \\
\ddot{l}= & \vec{a} \cdot \overrightarrow{n_{r}}-k_{r}^{2} \cdot \dot{s}^{2} \cdot \dot{l}  \\
l^{\prime \prime}= & \frac{\ddot{l}-l^{\prime} \cdot \ddot{s}}{\dot{s}^{2}}
\end{array}\right.
\end{equation}
\item Frenet to Cartesian \\
\begin{equation}
\left\{\begin{array}{lc}
x_{x}= & x_{r}-l \sin \left(\theta_{r}\right) \\
y_{x}= & y_{r}+l \cos \left(\theta_{r}\right) \\
\theta_{x}= & \arctan \left(\frac{l^{\prime}}{1-k_{r} l}\right)+\theta_{r} \in[-\pi, \pi] \\
v_{x}= & \sqrt{\left[\dot{s}\left(1-k_{r} l\right)\right]^{2}+\left(\dot{s} l^{\prime}\right)^{2}} \\
a_{x}= & \ddot{s} \frac{1-k_{r} l}{\cos \left(\theta_{x}-\theta_{r}\right)}+\frac{\dot{s}^{2}}{\cos \left(\theta_{x}-\theta_{r}\right)}\left[l^{\prime}\left(k_{x} \frac{1-k_{r} l}{\cos \left(\theta_{x}-\theta_{r}\right)}-k_{r}\right)-\left(k_{r}^{\prime} l+k_{r} l^{\prime}\right)\right] \\
k_{x}= & \left(\left(l^{\prime \prime}+\left(k_{r}^{\prime} l+k_{r} l^{\prime}\right) \tan \left(\theta_{x}-\theta_{r}\right)\right) \frac{\cos ^{2}\left(\theta_{x}-\theta_{r}\right)}{1-k_{r} l}+k_{r}\right) \frac{\cos \left(\theta_{x}-\theta_{r}\right)}{1-k_{r} l}
\end{array}\right.
\end{equation}

\end{itemize}

\subsection{Path Planning}
The objective of path planning is to offer a path with coordinates SL (Station and Lateral) so that our ego vehicle following this path will avoid collision with static and low-speed dynamic obstacles. As for the high-speed dynamic obstacles, the problem will be solved in the speed planning section. By merging the path profile from path planning and speed profile from speed planning, the final trajectory will be generated, which will satisfy our goals - avoid multiple dynamic and static obstacles. 
\subsubsection{Adaptive Sampling in SL graph}
\begin{figure}[!t]\centering
	\includegraphics[width=13.5cm]{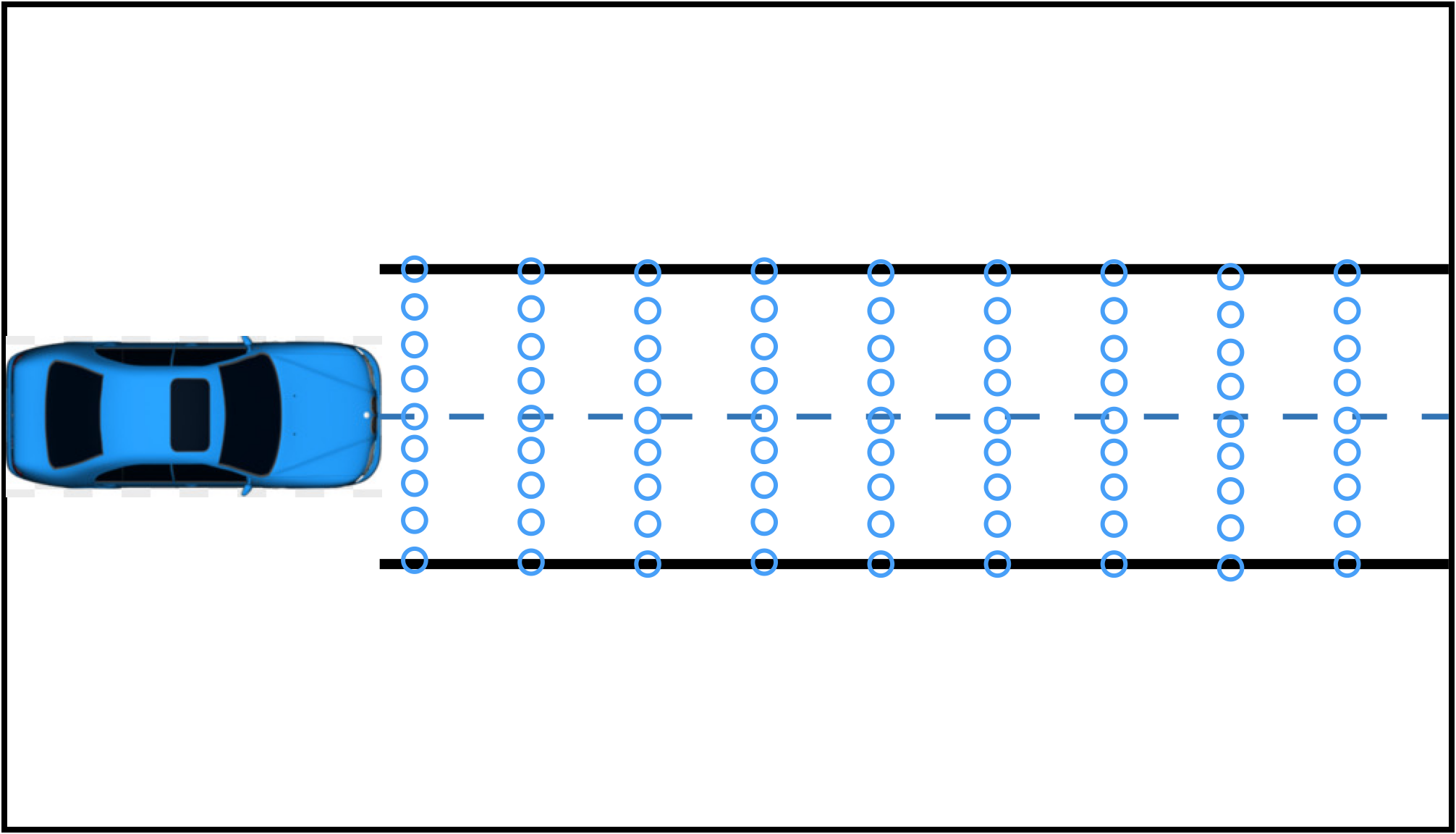}
	\caption{Average sampling with 9 rows and 9 columns }\label{average sampling}
	\end{figure}
In sampling-based motion planning algorithms, the number of samples determines the complexity of the computation required to solve the problem. Conventionally, the uniform sampling method will require a large number of samples and more computational resources as well. A uniform sampling pattern can be found in Gu et al. (2016), which is shown in the first picture of Fig \ref{average sampling}. The quality of the path generated by this uniform pattern is highly dependent on the resolution. However, more running time during dynamic programming in the next step is needed if we increased the resolution in uniform sampling to achieve a trajectory with higher performance. To solve the problem that existed in the uniform sampling method, an adaptive sampling method is proposed by us so as to dynamically adjust uniform sampling patterns based on the history information. Adaptive sampling will automatically adjust the density of sampling based on the partially-observed information and the historical experience in the real-time implementation. In our motion planning experiment, roads are always bounded by two edges, and it is easy to obtain the optimal trajectory from the last cycle and the obstacles information in the real-time implementation.  Therefore, we proposed our adaptive sampling method, which will take into consideration the historical trajectory, obstacles information and road boundaries. 

In the next, the details of the implementation of our Adaptive Sampling Method will be presented.
First of all, uniform sampling is realized as initialization, the same as shown in Figure \ref{average sampling}. Suppose the number of layers is $N$, and a layer is denoted by $L_i$ (i = 1,2,..., N), and in every layer, the number of nodes is M, so the location of a node on $L_i$ is denoted by $n_{ij}$ (i = 1,2,..., M). Then, inspired by Artificial Potential Field (APF), the repulsive field and attractive field are defined so as to predefine and compare the possibility of being nodes of optimal trajectory in the next cycle. 

Let us define the total APF force first, which equals the sum of repulsive force and attractive force.
In our problem, repulsive force is given by obstacles and road boundaries and the optimal path in the last cycle offers us the attractive field in the theory:

\begin{equation}
    U_{apf}(s, l) = U_{ref}(s, l) + U_{obs}(s, l) + U_{bound}(s, l)
\end{equation}

\begin{enumerate}
    \item [1)] Obstacle Repulsive Field \\
    The repulsive field given by static obstacles depends on the distance between sampling points and obstacles boundaries. Because the obstacle may not be symmetry with respect to the reference line, here we defined the obstacle field function based on the obstacle upper-bound $l_{obs\_up}(s)$ and obstacle lower-bound $l_{obs\_down}(s)$:

\begin{equation}
    U_{obs}(s,l) = \begin{cases}
\frac{1}{2}*\eta*\frac{1}{d^{2}(l, l_{obs\_up}(s) + 1)}, & l > l_{obs\_up(s) + 1} \\
    +inf, & l_{obs\_down(s)} - 1 \leq l \leq l_{obs\_up(s)} + 1 \\
    \frac{1}{2}*\eta*\frac{1}{d^2(l,l_{obs\_down(s)} - 1)}, & l < l_obs\_down(s) - 1
\end{cases}
\end{equation}

    Here $\eta$ is the coefficient of obstacle repulsive field. For the consideration of safety, it is defined as 1000 in our Simulink implementation.
    
    \item [2)] Boundary Repulsive Field \\
    Then, the boundary repulsive field is given here. Similar to the form of the obstacle field, the form of the equation is defined depending on the distance between lateral deviation and lateral road upper boundary $l_{bound\_up}(s)$, the distance between that and lateral road lower boundary
$l_{lbound}(s)$ as well:
\begin{equation}
    U_{bound}(s,l) = \begin{cases}
+inf, & l \geq l_{ubound}(s) - 1 \\
\frac{1}{2}*\gamma*(\frac{1}{d^2(l, l_{ubound}(s) - 1} + \frac{1}{d^2(l, l_{lbound}(s) + 1)}, & l_{lbound}(s) + 1 < l < l_{ubound}(s) -1 \\
+inf, & l \leq l_{lbound}(s) + 1
\end{cases}
\end{equation}

$\gamma$ is the coefficient of the boundary repulsive field. By testing in our experiment, if the value is chosen as 20, our adaptive sampling method performs well. In addition, it is important to mention that, the objective of forming the equation when $l_{lbound}(s) + 1 < l < l_{ubound}(s) + 1$ like that is to make sure the continuity of equations.

\item [3)] Trajectory Attractive Field \\
In a real implementation, it is not difficult to obtain the optimal trajectory in the last cycle of motion planning. This information can be fully used to converge the unique sampling area into a more feasible and smaller space for the reason that the optimal trajectory will not vibrate too much in 100 milliseconds time intervals in a normal situation.

\begin{flushleft}
$$U_{ref}(s, l) = \frac{1}{2}*\alpha*d^2(l, l_{ref}(s))$$
\end{flushleft}

Here $\gamma$ is the coefficient of the trajectory attractive field. In our experiment, the value is chosen as 20. In the first cycle, there is no optimal solution from the last cycle, which will lead to the deadlock problem in the Simulink simulation. To solve this problem, a delay with initial values is added to this signal. The initial value is given from the reference line.
\end{enumerate}

\begin{figure}[!t]\centering
	\includegraphics[width=13.5cm]{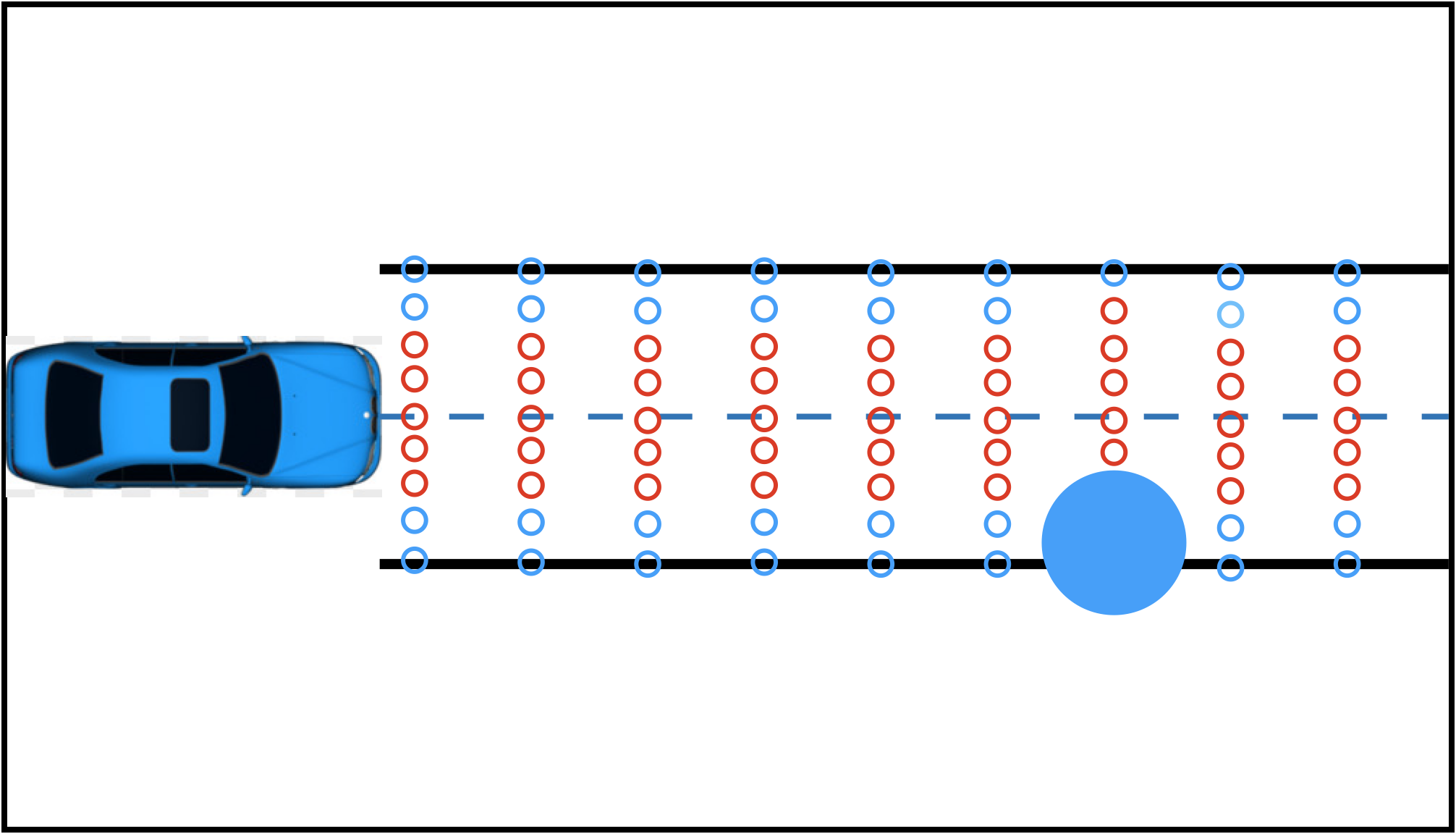}
	\caption{Adaptive samplings with obstacles }\label{Adaptive samplings with obstacles}
	\end{figure}

In this Figure \ref{Adaptive samplings with obstacles}, we see the visualization of the effect of our ASAPF. Depending on the distance with obstacles, road boundaries and history optimal trajectory, the sampling points in each column are adaptively adjusted so as to focus more on the area where the next optimal trajectory is easier to appear.

\subsubsection{Dynamic Programming in SL graph}
Until now, we have already transformed the reference line generated based on the global route from Cartesian into Frenet coordinate system, and our proposed adaptive sampling method has already been implemented to decrease the scale of sampling meanwhile maintaining the possibility to obtain optimal path. Then, in this section, the problem becomes how to find an optimal function of lateral coordinate $l = f(s)$ w.r.t. station coordinate in non-convex SL space. Quadratic programming is fast and efficient enough to solve a 2-D optimization problem but the premise is that the problem has to be within convex space. It is obvious that our motion planning task is not a convex problem. Therefore, dynamic programming is used before quadratic programming to make path decisions, or in other words, provide a rough path profile. Then based on that rough path profile, a convex space will be built where quadratic programming could be implemented easily to take into consideration dynamic vehicle constraints and make the final path smoothing which is important for autonomous vehicles to follow in the control part.

After constructing sampling space using the adaptive sampling method, each graph edge could be evaluated by the summation of cost functions. Smoothness, obstacle avoidance and reference line should be considered together in dynamic programming. Therefore, the total cost of every edge in dynamic programming could be defined in the form below:
\begin{equation}
     C_{total}(f(s)) = C_{smooth}(f(s)) + C_{obs}(f(s)) + C_{ref}(f(s))
\end{equation}

\begin{enumerate}
    \item [1)] Smoothing Cost \\
    Let us give the smoothing cost first. By Newton's theory, it is easy to know that lateral velocity $f^{'}(s)$ is the derivative of a lateral function $f(s)$. Due to the same reason, the derivative of lateral velocity $f^{'}(s)$ means lateral acceleration $f^{''}(s)$. Here lateral jerk $f^{'''}(s)$ is defined as the derivative of lateral acceleration $f^{''}(s)$, which represents the rate at which an object's acceleration changes with respect to time. The jerk will influence the comfort of passengers, so it should be decreased as possible as we can. The smoothing cost has the form below:
    \begin{equation}
        C_{smooth}(f(s)) = w_1\int (f^{'}(s))^2ds + w_2\int (f^{''}(s))^2ds + w_3\int (f^{'''}(s))^2ds
    \end{equation}
    
    \item [2)] Obstacle Cost \\
    If the distance between lateral derivation and the obstacle centre point is less than the boundary, it means there will be a collision at that point, so a large value will be given as obstacle cost so as to detect an infeasible path. Here in equations, for simplicity of expression, we define the large value as $+inf$. Then, if the distance is over the safety distance, the influence of that obstacle could be neglected, so the obstacle cost will become 0. Between collision distance and safety distance, the obstacle cost will be defined as a monotonically decreasing function with respect to the distance, as shown below:
    
    \begin{equation}
        C_{obs}(f(s)) = \begin{cases}
        0, & f(s) > d_{safety} \\
        w*\frac{1}{d^2(f(s),  d_{collision})}, & d_{collision} < f(s) < d_{safety} \\
        +inf, & f(s) < d_{collision}
        \end{cases}
    \end{equation}
    
    \item [3)] Reference Cost \\
    If there are no obstacles on the road, our ego vehicle should try to follow our reference line. Therefore, a reference cost is defined to incrementally punish the path with respect to the distance with the reference line: 
    \begin{equation}
        C_{ref}(f(s)) = \int (f(s) - l_{ref}(s))^2 ds
    \end{equation}

\end{enumerate}

In the content above, the cost function is defined which could give cost to every single edge. We could set the initialization point as layer 0, and the cost of nodes in this layer is 0 too. The costs of nodes in the current layer are equal to the minimal sum of edge cost and cost of nodes in the last layer. Repeat that step until the nodes in the final layer are given values of costs as well. Then try to search the minimal cost in each layer from the final layer back to layer 0, which will give us a feasible path that is the result of dynamic programming. The details of the DP process are given in Algorithm \ref{Dynamic Programming}.
\begin{algorithm}
    \caption{Dynamic Programming}
    \label{Dynamic Programming}
    \begin{algorithmic}
        \STATE $col = {1,...,N}$
        \STATE $row = {1,...,M}$
        \STATE $Cost_{0,0} = 0$
         \STATE $Cost_{i,j} = +inf$
        \FOR{$i \in col, j \in row$}

        \FOR{$k \in row$}
        \STATE $Cost_{edge}(k,j) = C_{smooth}(k,j) + C_{obs}(k,j) + C_{ref}(k,j)$
        \STATE $Cost(i,j) = Cost_{i-1, k} + Cost_{edge}(k,j)$
        \IF{$Cost(i,j) < Cost_{i,j}$}
            \STATE $Cost_{i,j} = Cost(i,j)$
        \ENDIF
        \ENDFOR
        \ENDFOR

    \end{algorithmic}
\end{algorithm}

In our implementation, it is easy to know that it is impossible to connect sampled nodes using straight lines because they can not satisfy the constraints of curvature and jerk. To solve this problem, the fifth-degree polynomial curve is used which could satisfy the requirement to jerk and does not make the problem too complex to solve at the same time. In Appendix \ref{Appendix 1}, I will explain why the fifth-degree is chosen, rather than another degree, for example, the fourth-degree polynomial which is easier to solve, and the sixth-degree polynomial which is more accurate.

\subsubsection{Quadratic Programming in SL graph}
A rough path profile has already been obtained by dynamic programming (QP) in the last step. That means we have already decided to pass the obstacle from the left side or right side. Next, a convex space is built based on the obtained rough path. Quadratic Programming could be seen as a refinement of the dynamic programming path step. Normally, before implementing QP, objective functions and constraints should be defined in advance. The objective function is mainly related to a linear combination of smoothness costs which have the same form as in dynamic programming. Mathematically, the QP path step will optimize the following function:
\begin{equation}
\label{objective function}
    C_s(f) = w_1\int (f^{'}(s))^2ds + w_2\int (f^{''}(s))^2ds + w_3\int (f^{'''}(s))^2ds
       + w_4\int (f(s) - r(s))^2 ds
\end{equation}

Here $r(s)$ represents the DP path result. 

The constraints in the QP path include boundary constraints and dynamic feasibility. They are defined in discrete formulation, so let us note:
\begin{equation}
    l_i = f(s_i), l_i^{'} = f^{'}(s_i), l^{''} = f^{''}(s_i)
\end{equation}
Suppose the third order derivative of $f(s)$ between $l_i$ and $l_{i+1}$ is a constant and the higher orders of that are 0. So we could use Taylor expansion here in $l_i$ and $l_{i+1}$.
\begin{equation}
    \begin{aligned}
&l_{i+1}=l_{i}+l_{i}^{\prime} \cdot \Delta s+\frac{1}{2} l_{i}^{\prime \prime} \cdot \Delta s^{2}+\frac{1}{6}\left(\frac{l_{i+1}^{\prime \prime}-l_{i}^{\prime \prime}}{\Delta s}\right) \Delta s^{3} \\
&l_{i+1}^{\prime}=l_{i}^{\prime}+l_{i}^{\prime \prime} \cdot s+\frac{1}{2}\left(\frac{l_{i+1}^{\prime \prime}-l_{i}^{\prime \prime}}{\Delta s}\right) \cdot \Delta s^{2} 
\end{aligned}
\end{equation}
By simple transformation for that equation, we will have:
\begin{equation}
    \begin{aligned}
    &l_{i}+\Delta s \cdot l_{1}^{\prime}+\frac{1}{3} \Delta s^{2} \cdot l_{i}^{\prime \prime}-l_{i+1}+\frac{1}{6} \Delta s^{2} \cdot l_{i+1}^{\prime \prime}=0 \\
&l_{1}^{\prime}+\frac{1}{2} \Delta s \cdot l_{i}^{\prime \prime}-l_{i+1}^{\prime}+\frac{1}{2} \Delta s \cdot l_{i+1}^{\prime \prime}=0
    \end{aligned}
\end{equation}
It could be written in matrix form:
\begin{equation}
\label{equation constraints}
\left(\begin{array}{cccccc}
1 & \Delta s & \frac{1}{3} \Delta s^{2} & -1 & 0 & \frac{1}{6} \Delta s^{2} \\
0 & 1 & \frac{1}{2} \Delta s & 0 & -1 & \frac{1}{2} \Delta s
\end{array}\right)\left(\begin{array}{l}
l_{i} \\
l_{i}^{\prime} \\
l_{i}^{\prime \prime} \\
l_{i+1} \\
l_{i+1}^{\prime} \\
l_{i+1}^{\prime \prime}
\end{array}\right)=\left(\begin{array}{l}
0 \\
0
\end{array}\right)
\end{equation}
Note the left matrix on the left side of the above equation is $A_{eq\_sub}$. This constraint is only for two neighbour points. Suppose we have $l_1,l_1^{\prime}, l_1^{\prime \prime}, ... , l_n, l_n^{'}, l_n^{''}$ that needs to be optimized. If we implement this constraint for each two neighbour points, we will have the total equation constraint:
$A_{eq}x = b_{eq}$

Then the simple inequality constraint could be defined as:
\begin{equation}
\left(\begin{array}{c}
\operatorname{l}_{min1} \\
-\infty \\
-\infty \\
\operatorname{l}_{min2} \\
-\infty \\
-\infty \\
\vdots
\end{array}\right) \leq x \leq \left(\begin{array}{c}
\operatorname{l}_{max1} \\
+\infty \\
+\infty \\
\operatorname{l}_{max2} \\
+\infty \\
+\infty \\
\vdots
\end{array}\right)
\end{equation}
\begin{figure}[!t]\centering
	\includegraphics[width=13.5cm]{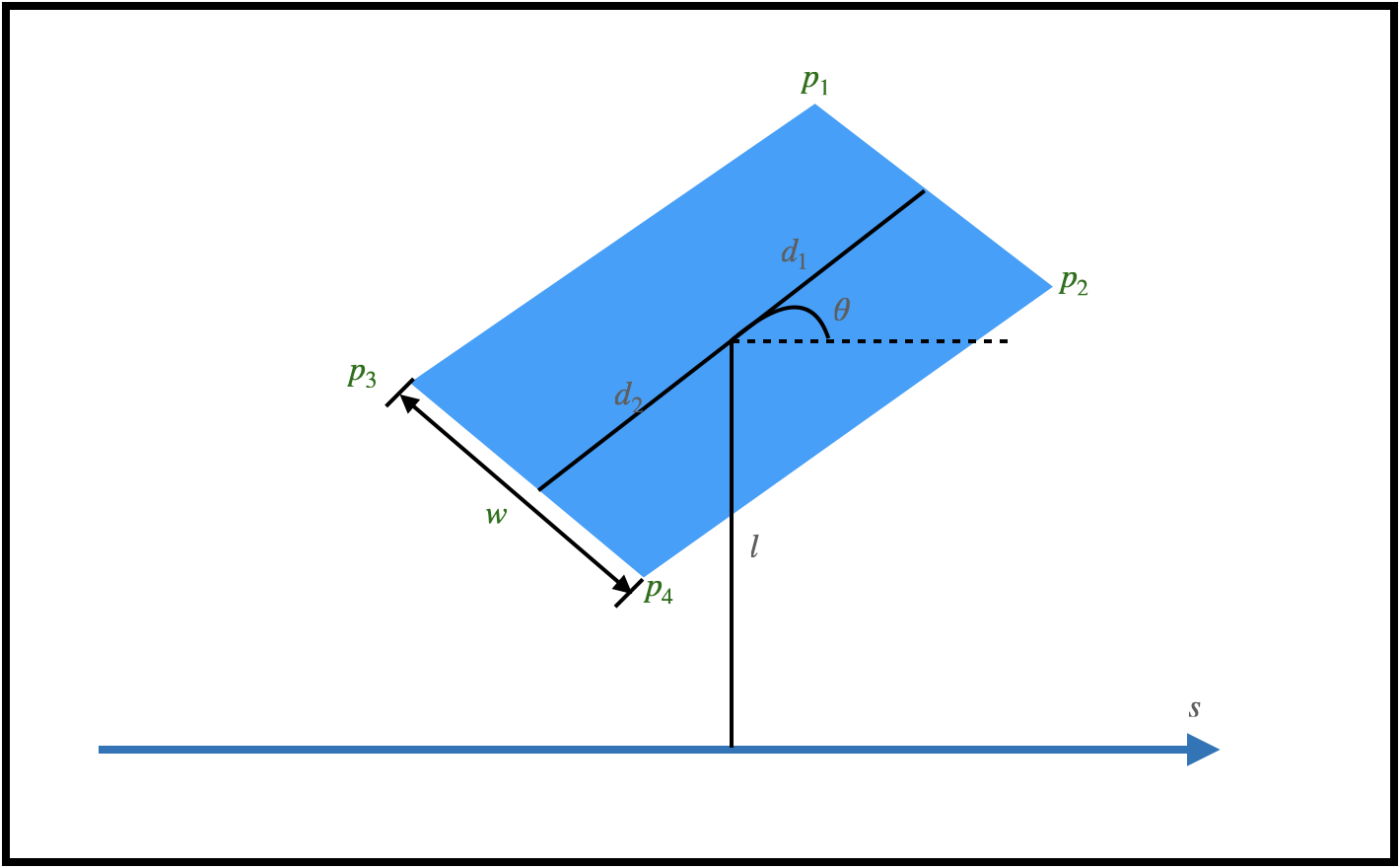}
	\caption{Vehicle volume constraints }\label{vehicle volume constraints}
	\end{figure}
If we consider the volume of the vehicle, like in Figure \ref{vehicle volume constraints}, we need to make sure that all four edge points have to be within the range of the road boundary. Therefore, we have the expression of four endpoints:
\begin{equation}
\label{four end points}
\begin{aligned}
&l_{p1}=l+d_{1} \cdot \sin \theta+\frac{w}{2} \cdot \cos \theta \\
&l_{p2}+l+d_{1} \cdot \sin \theta-\frac{w}{2} \cos \theta \\
&L_{p3}=l-d_{2} \cdot \sin \theta+\frac{w}{2} \cos \theta \\
&l_{p4}=l-d_{2} \cdot \sin \theta-\frac{w}{2} \cdot \cos \theta
\end{aligned}
\end{equation}
Because $\theta$ is a very small value, by trigonometric functions, it is easier to do the approximation:
\begin{equation}
    sin \theta \approx tan \theta \approx l^{\prime}, cos \theta \approx 1
\end{equation}
So the equation \ref{four end points} could be rewritten as below:
\begin{equation}
\left\{\begin{array}{l}
l_{p_{1}}=l_{i}+d_{1} \cdot l_{i}^{\prime}+\frac{w}{2} \in\left(l_{\text {mini }} \operatorname{l}_{maxi}\right) \\
l_{p_{2}}=l_{i}+d_{1} \cdot l_{i}^{\prime}-\frac{w}{2} \in\left(l_{\text {mini }}, l_{\text {maxi }}\right) \\
l_{p_{3}}=l_{i}-d_{2} \cdot l_{i}^{\prime}+\frac{w}{2} \in\left(l_{\text {mini }}, l_{\text {maxi }}\right) \\
l_{p 4}=l_{i}-d_{2} \cdot l_{i}^{\prime}-\frac{w}{2} \in\left(l_{\text {mini }}, l_{\text {maxi }}\right)
\end{array}\right.
\end{equation}

For safer manoeuvre and easier implementation, we first find the minimum of $l_maxi$ between $s_i - d_2$ and $s_i + d_1$ and note it as $ub_i$. So our constraints become:
\begin{equation}
\left\{\begin{array}{l}
l_{p1} \leq u b_{i} \\
l_{p2} \leq u b_{i} \\
l_{p3} \leq u b_{i} \\
l_{p4} \leq u b_{i}
\end{array}\right.
\end{equation}

For the same reason, the same constraints are implemented for the lower boundary, therefore, in the end, we will have the inequality constraint:
\begin{equation}
\label{inequality}
\left\{\begin{array}{l}
l b_{i} \leq l_i + d_{1} \cdot l_{i}^{\prime}+\frac{w}{2} \leq u b_{i} \\
l b_{i} \leq l_{i}+d_{1} \cdot l_{i}^{\prime} -\frac{w}{2} \leq u b_{i} \\
l b_{i} \leq l_{i} - d_{2} \cdot l_{i}^{\prime} + \frac{w}{2} \leq u b_{i} \\
l b_{i} \leq l_{i} - d_{2} \cdot l_{i}^{\prime} - \frac{w}{2} \leq u b_{i}
\end{array}\right.
\end{equation}
Rewrite \ref{inequality} in matrix form, we will obtain:
,
\begin{equation}
\left(\begin{array}{l}
l b_{i}-\frac{\omega}{2} \\
l b_{i}+\frac{\omega}{2} \\
l b_{i}-\frac{\omega}{2} \\
l b_{i}+\frac{\omega}{2}
\end{array}\right) \leq \left(\begin{array}{ccc}
1 & d_{1} & 0 \\
1 & d_{1} & 0 \\
1 & -d_{2} & 0 \\
1 & -d_{2} & 0
\end{array}\right)\left(\begin{array}{l}
l_{i} \\
l_{i}^{\prime} \\
l_{i}^{\prime \prime}
\end{array}\right) \leq \left(\begin{array}{c}
u b_{i}-\frac{\omega}{2} \\
u b_{i}+\frac{\omega}{2} \\
u b_{i}-\frac{\omega}{2} \\
u b_{i}+\frac{\omega}{2}
\end{array}\right)
\end{equation}
Then the last step is to write it in $A_{sub}x \leq b_{sub}$ to satisfy the inequality constraint formulation in Quadratic Programming:
\begin{equation}
\left(\begin{array}{ccc}
1 & d_1 & 0 \\
1 & d_1 & 0 \\
1 & -d_2 & 0 \\
1 & -d_{2} & 0 \\
-1 & -d_1 & 0 \\
-1 & -d_1 & 0 \\
-1 & d_{2} & 0 \\
-1 & d_2 & 0
\end{array}\right) 
\cdot\left(\begin{array}{l}
l_{i} \\
l_{i}^{\prime} \\
l_{i}^{\prime \prime}
\end{array}\right)
\leq \left(\begin{array}{l}
u b_{i}-\frac{\omega}{2} \\
u b_{i}+\frac{\omega}{2} \\
u b_{i}-\frac{\omega}{2} \\
u b_{i}+\frac{\omega}{2} \\
-l b_{i}+\frac{\omega}{2} \\
-l b_{i}-\frac{\omega}{2} \\
-l b_{i}+\frac{\omega}{2} \\
-l b_{i}-\frac{\omega}{2}
\end{array}\right)
\end{equation}

Until now, we have already given the Objective Function \ref{objective function}, Equation Constraint \ref{equation constraints} and Inequality Constraint \ref{inequality} in Quadratic Programming. It is easier to solve the optimization problem with this formulation in Matlab/Simulink.
\subsection{Speed Planning}
As for static obstacles and dynamic obstacles at low speeds, the path profile obtained by path planning is enough to realize obstacle avoidance. However, in automated driving motion planning problems, it is totally not sufficient if only static and low-speed obstacles are considered for the reason that there are always other vehicles, pedestrians, bicycles, etc on the road. Therefore, speed planning is designed to avoid dynamic obstacles. We admitted that path planning has already been finished. That means we have a feasible path profile to avoid obstacles. Based on that path profile, speed planning will tell the automated vehicle to drive following a specific velocity so as to avoid dynamic obstacles. Similar to path planning, the first step is also adaptive sampling, but in the ST graph. Then, the formulations of DP and QP for the ST graph in adaptive sampling space are analysed.  
\subsubsection{Adaptive Sampling in ST graph}
Firstly, the station coordinate system is the path in a Cartesian coordinate system. In the next several seconds, if the dynamic obstacles will be projected on the station coordinate system, that will be presented in the ST graph. 

In the past, uniform sampling is implemented in the ST graph, leading to huge computation complexity. Therefore, we also proposed to implement adaptive sampling using dynamic vehicle constraints here so as to delete some impossible nodes before dynamic programming. 
To formalize the problem, vehicle dynamics are given in the form:
$v \in [v_{min}, v_{max}]$, $a \in [a_{min}, a_{max}]$, $J \in [J_{min}, J_{max}]$

Similar to path planning, here we defined the cost function of our ASAPF:
\begin{equation}
    U_{total}(s(t)) = U_{smooth}(s(t)) + U_{obs}(s(t)) + U_{constraint}(s(t))
\end{equation}

\begin{enumerate}
    \item [1)] Smoothing Cost
    \begin{equation}
        U_{smooth}(s, t) = w_1 \int (s^{'}(t) - s^{'}(t-1))^2dt
+ w_2 \int (s^{''}(t))dt + w_3 \int (s^{'''}(t))dt
    \end{equation}

    \item [2)] Obstacle Cost
    \begin{equation}
        U_{obs}(s, t) = \begin{cases}
    \frac{1}{2}*w*\frac{1}{d^{2}(s(t), s_{obs\_upper}(t) + 1)}, & s(t) > s_{obs\_upper}(t) + 1 \\
    +inf, & s_{obs\_lower}(t) - 1 \leq l \leq s_{obs\_upper}(t) + 1 \\
    \frac{1}{2}*w*\frac{1}{d^2(s(t),s_{obs\_lower(t)} - 1)}, & s(t) < s_obs\_lower(t) - 1
    \end{cases}
    \end{equation}
    \item [3)] Constraint Cost
    \begin{equation}
        U_{constraint}(s, t) = \begin{cases} 
        0, & s(t) - s(t-1) \in [v_{min}, v_{max}] \\
        & and \ s^{'}(t) - s^{'}(t-1) \in [a_{min}, a_{max}] \\
        & and \ s^{''}(t) - s^{''}(t-1) \in [J_{min}, J_{max}] \\
        +inf, & otherwise
    \end{cases}
    \end{equation}
    
\end{enumerate}
\begin{figure}[!t]\centering
	\includegraphics[width=13.5cm]{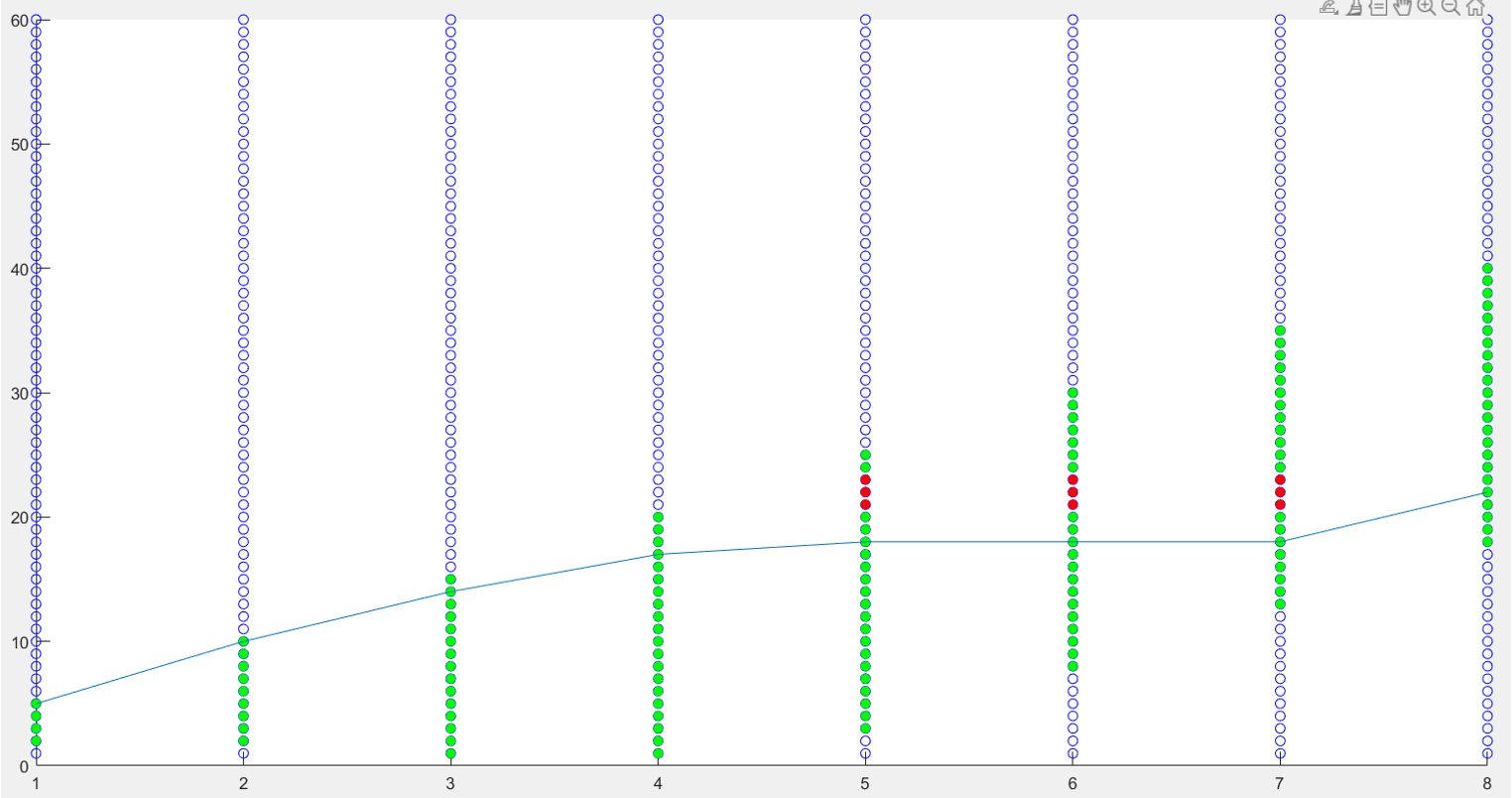}
	\caption{Adaptive sampling in ST graph}\label{Adaptive sampling in ST graph}
	\end{figure}
For example, if $v_0 = 0 m/s$, $a_0 = 0$, prediction period $T = 8s$, $v \in [-5, 5]$, $a \in [-2, 2]$, by implementing our adaptive sampling method in ST graph, we will have Figure \ref{Adaptive sampling in ST graph}. During this figure, the blue points are the result of average sampling, and the green points are the result of our adaptive sampling. We could see from the graph that the useless nodes have already been removed before Dynamic Programming in ST graph, therefore the computational task is reduced too, same as adaptive sampling in path planning.

\subsubsection{Dynamic Programming in ST graph}
From the perception module, it is not difficult to obtain information on dynamic obstacles (x coordinate, y coordinate, velocity and heading angle). Then, these dynamic obstacles need to be projected on the ST graph we built in the last step to represent the prediction of the trajectory of dynamic obstacles. For the reason of simplicity, the velocity of dynamic obstacles is supposed to be constant. This assumption will not lead to the failure of speed planning, because our planning time interval is very short (around 100 milliseconds). During this short interval, speed could be approximated as constant. Similar to path planning, the goal of speed planning is to find the best speed profile on the ST graph which is a non-convex optimization problem too. Therefore, dynamic programming combined with quadratic programming is used to find a smooth speed profile on the ST graph. In detail, a piece-wise linear speed profile function is represented as $s = (s_0, s_1, ..., s_n)$ on the grids. The finite difference method is taken to approximate the derivatives. If $dt$ represents the time interval on the time axis between two evaluated points with equal space gap, then it is easy to give the formulation of velocity $v_i$, acceleration $a_i$, and jerk $j_i$:
\begin{equation}
    v_i = s^{'}_i \approx \frac{s_i - s_{i-1}}{dt} 
\end{equation}

\begin{equation}
    a_i = s^{''}_i \approx \frac{s_i - 2s_{i-1} + s_{i-2}}{(dt)^2} 
\end{equation}
\begin{equation}
    j_i = s^{'''}_i \approx \frac{s_i - 3s_{i-1} - 3s_{i-2} + s_{i-3}}{(dt)^3} 
\end{equation}

The cost function needs to be specifically defined for the speed profile here because it is different from that in the SL graph. The idea is also to optimize a cost function in the ST graph within the constraints. In detail, the cost function for dynamic programming in speed planning is represented as follows:
\begin{equation}
    C_{total}(s) = w_1\int_{t_0}^{t_n}g(s' - v_{ref})dt
+ w_2\int_{t_0}^{t_n}((s^{''})^2)dt + w_3\int \int_{t_0}^{t_n}(s_{'''})^2dt + w_4*C_{obs}(s)
\end{equation}

If there are no obstacles, the vehicle should try to follow the reference velocity, so the first term is the velocity-keeping cost. The acceleration and jerk square integral describes the smoothness of the speed profile. The last term, $C_{obs}$ decided by the distance with obstacles' boundaries, describes the total obstacle cost. 

\subsubsection{Quadratic Programming in ST graph}
After Dynamic Programming in the last step, we have already known that the autonomous vehicle needs to accelerate or decelerate to avoid dynamic obstacles avoidance. Then, the next step is to refine the speed profile so as to improve the success of path following in the Control module and the comfort of passengers. Same as in path planning, we defined the cost function and constraints in Quadratic Programming too here.
The cost function is described as follows:
\begin{equation}
  C_{total}(s) = w_1\int_{t_0}^{t_n}(s - s_{ref})^2dt
+ w_2\int_{t_0}^{t_n}((s^{''})^2)dt + w_3\int \int_{t_0}^{t_n}(s_{'''})^2dt 
\end{equation}

The constraints of QP:
$$s(t_i) \leq s(t_{i+1}), i = 0,1,2,...,n-1, $$
$$s_{lower}(t_i) \leq s(t_i) \leq s_{upper}(t_i), $$
$$v(t_i) \leq V_{upper},$$
$$A_{min} \leq a(t_i) \leq A_{max} ,$$
$$J_{min} \leq j(t_i) \leq J_{max}$$
which is given in \cite{fan2018baidu}. 

\newpage
\section{Vehicle Models}
\label{Vehicle Models}

\subsection{Kinematic Model}

According to \cite{samak2021control}, the Kinematic model of autonomous vehicles is the study of the motion of a system without taking into consideration the forces and torques. It is sufficiently approximate to perform non-aggressive manoeuvres at lower speeds. However, if we consider something like a racing car, automated driving on the highway or in complex scenarios, the kinematic model would fail to capture the actual system dynamics. In such occasions, dynamic models are needed to consider forces and torques as well. 

In this section, the kinematic bicycle model which is the most widely used nowadays is presented. In practice, this model tends to strike a good balance between simplicity and accuracy and is therefore widely adopted. The basic idea of modelling is to see how the vehicle states change over time based on the previous states and current control inputs. First, vehicle states are defined, constituting $x$ and $y$ components of position, heading angle or orientation $\theta$ and velocity $v$. In other words, self-vehicle state vector $q$ is defined as follows:
\begin{equation}
q = [x, y, \theta, v]^T
\end{equation}
First, control vector $u$ is defined as following: 
\begin{equation}
u = [a, \delta]^T
\end{equation}
Due to the limitation of acceleration and deceleration based on the physical throttle and braking, the input $a$ should be limited in the range of $[-a'_{max}, a_{max}]$. Additionally, The steering command alters the steering angle $\delta$ of the vehicle, where $\delta \in [-\delta_{max}, \delta_{max}]$ such that negative steering angles dictate left turns and positive steering angles otherwise.
\begin{figure}[!t]\centering
	\includegraphics[width=13.5cm]{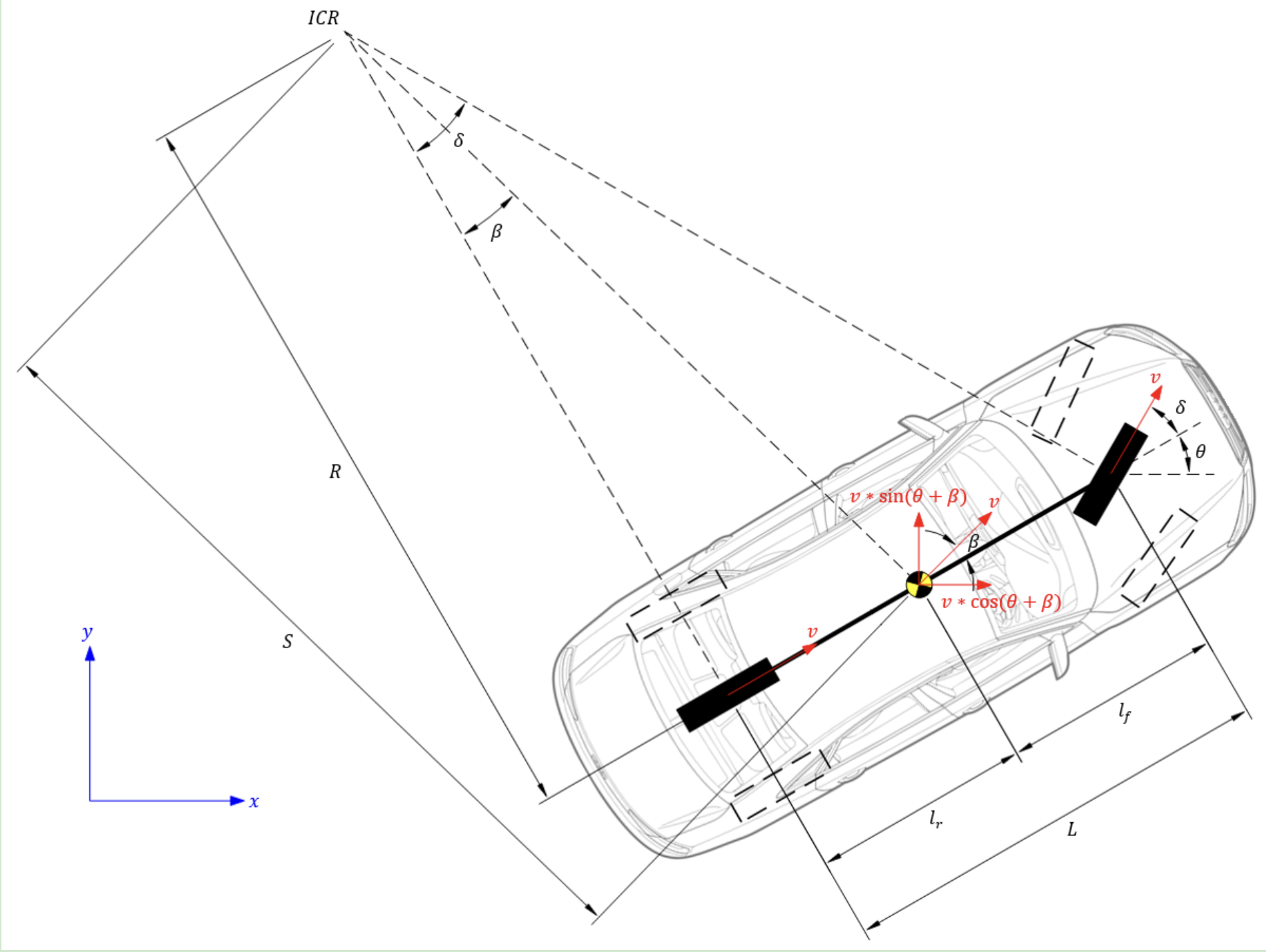}
	\caption{Kinematic model}\label{kinematic model}
\end{figure}
The kinematic model of the vehicle as shown in figure \ref{kinematic model}. The slip angle $\beta$ can be computed using the distance between the rear wheel axle and the vehicle's centre gravity, such that:
\begin{equation}
    tan(\beta) = \frac{l_r}{R} = \frac{l_r}{\frac{L}{tan(\delta)}} = \frac{l_r}{L}*tan(\delta)
\end{equation}  
therefore,
\begin{equation}
    \beta = tan^{-1}(\frac{l_r}{L}*tan(\delta))
\end{equation}
The form of the slip angle could be further simplified if we ideally choose $l_r$ as half of L. Then we have: 
\begin{equation}
    \beta = \tan^{-1}(\frac{\tan\delta}{2})
\end{equation}
The derivative of x and y is equal to the vectors following the direction of coordinate x and coordinate y separately. Therefore, we could easily compute them with the relation of $\theta$, $\beta$ and velocity vector $v$.
\begin{equation}
    \Dot{x} = \Dot{v}\cos(\theta + \beta)
\end{equation}
\begin{equation}
    \Dot{y} = \Dot{v}\sin(\theta + \beta)
\end{equation}
The next step is to calculate $\Dot{\theta}$. According to Figure \ref{kinematic model},
\begin{equation}
    \Dot{\theta} = \frac{v}{S}
\end{equation}
So $S$ should be computed in advance. Following the basic trigonometry, it is easy to get:
\begin{equation}
    S = \frac{R}{\cos\beta}
\end{equation}
In addition, 
\begin{equation}
    R = \frac{L}{\tan\delta}
\end{equation}
Therefore,
\begin{equation}
    S = \frac{L}{\tan\delta\cos\beta}
\end{equation}
Until now, we could combine 0.8 and 0.11 to compute $\Dot{\theta}$, and
\begin{equation}
    \Dot{\theta} = \frac{v\tan\delta\cos\beta}{L}
\end{equation}
Finally, we can compute $\Dot{v}$ using differential relation.
\begin{equation}
    \Dot{v} = a
\end{equation}
Now let us formulate the continuous-time kinematic model of autonomous vehicle
\begin{equation}
    \Dot{q} = \begin{bmatrix}
                \Dot{x}\\
                \Dot{y}\\
                \Dot{\theta}\\
                \Dot{v}
                \end{bmatrix}
            = \begin{bmatrix}
                v\cos(\theta + \beta)\\
                v\sin(\theta + \beta)\\
     \frac{v\tan\delta\cos\beta}{L}\\
                a
                \end{bmatrix}
\end{equation}
The last step is to transfer the kinematic model from continuous time to discrete time to obtain a state transition equation($q_{t+1} = q_t + \Dot{q_t}\Delta t$). Here $t$ means the current time instant and $t + 1$ represents the next time instant.

$$\left\{
\begin{aligned}
    x_{t+1} = x_t + \dot x_t\Delta t\\
    y_{t+1} = y_t + \dot y_t\Delta t\\
    \theta_{t+1} = \theta_t + \dot\theta d\Delta t\\
    v_{t+1} = v_t + \dot v_t\Delta t
\end{aligned}\right.
$$

\subsection{Dynamic Model}
\begin{figure}[!t]\centering
	\includegraphics[width=13.5cm]{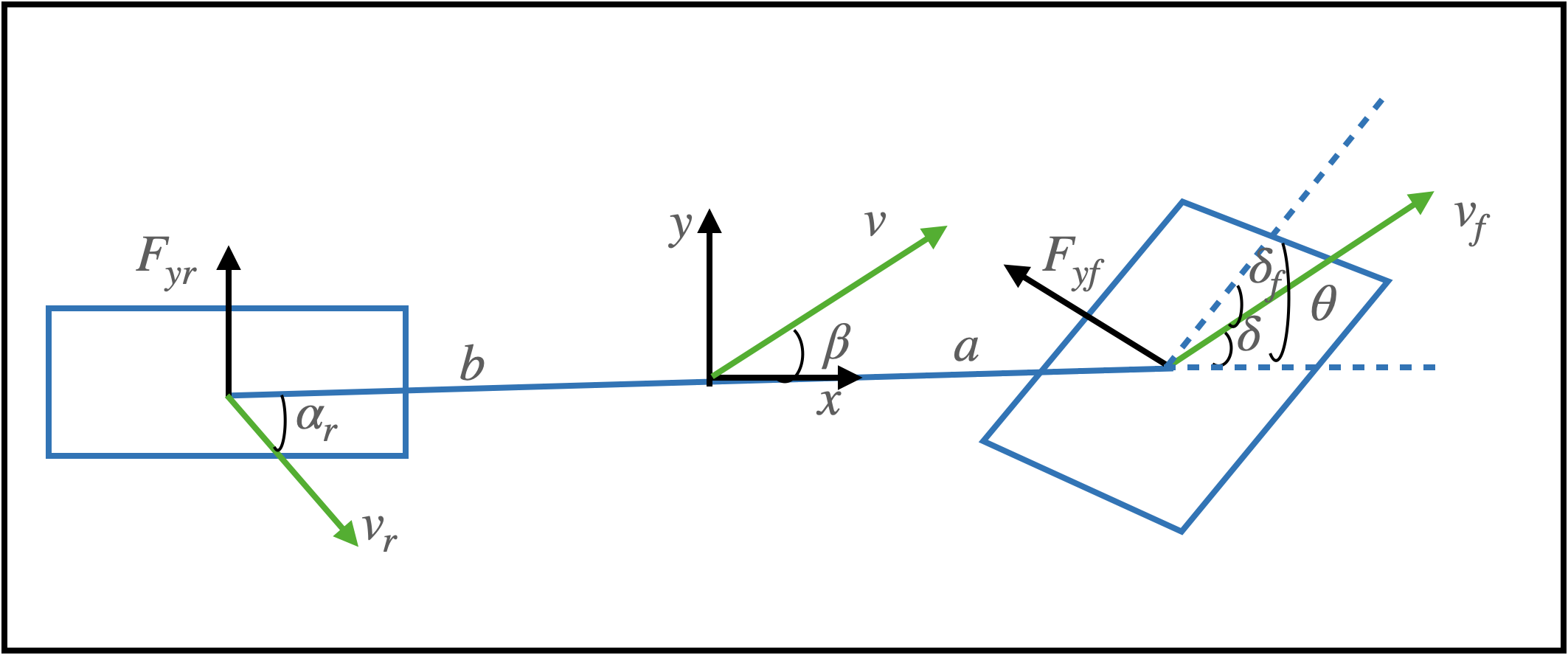}
	\caption{Dynamic model of 2-D bicycle vehicle }\label{Dynamic model of 2-D bicycle vehicle}
\end{figure}
The dynamic model of two wheels in a 2-D bicycle vehicle is shown in Figure \ref{Dynamic model of 2-D bicycle vehicle}, based on which we will analyze the dynamic model of vehicles. Suppose the front wheel angle $\delta$ is small in our model, and the lateral force of the wheels follows the equation below (C means cornering stiffness):
\begin{equation}
    F = C \cdot \alpha
\end{equation}
Then, by Newton's Law, we have:
\begin{equation}
\label{newton's law}
\begin{aligned}
& m a_y =F_{yf} \cdot \cos \delta +F_{yr} \\
& F_{y f} \cdot \cos \delta \cdot a - F_{y r} \cdot b = I \cdot \ddot{\varphi}
\end{aligned}
\end{equation}
Because $\delta$ is very small, $cos \delta \approx 1$, so the function \ref{newton's law} will become:
\begin{equation}
\label{dynamic equations}
\begin{aligned}
 & m \cdot a_{y}=F_{yf}+F_{yr}=C_{\alpha f} \cdot \alpha_{f}+C_{\alpha f} \cdot \alpha_{r} \\
& I \cdot \ddot{\varphi}=F_{yf} \cdot a-F_{yr} \cdot b=a \cdot C_{\alpha f} \cdot \alpha_{f}-b \cdot c_{\alpha r} \cdot \alpha_r 
\end{aligned}
\end{equation}
From Figure \ref{Dynamic model of 2-D bicycle vehicle}, it is easy to obtain the expression of $a_y$, $\alpha_f$ and $\alpha_r$.
\begin{equation}
\left\{\begin{array}{l}
a_{y}=\ddot{y}+v_{x} \cdot \dot{\varphi} \\
\alpha_{y}=\frac{v_{y}-\dot{\varphi} b}{v_{x}} \\
\alpha_{f}=\frac{\dot{\varphi} \cdot a+v_{y}}{V_{x}}-\delta
\end{array}\right.
\end{equation}
So Equation \ref{dynamic equations} could be expressed as:
\begin{equation}
\left\{\begin{array}{l}
m(\ddot{y}+v_x \cdot \dot{\varphi})=C_{\alpha f}\left(\frac{\dot{\varphi} a+v_y}{v_x}-\delta\right)+C_{\alpha r} \cdot\left(\frac{v_{y}-\dot{\varphi} \cdot b}{v_{x}}\right) \\
I \cdot \ddot{\varphi}=a \cdot C_{\alpha f} \cdot\left(\frac{\dot{\varphi} a+v_y}{v_{x}}-\delta\right)-b \cdot C_{\alpha r} \cdot\left(\frac{v_{y}-\dot{\varphi} b}{v_{x}}\right)
\end{array}\right.
\end{equation}
Then it could be written in matrix form:
\begin{equation}
\left(\begin{array}{c}
\ddot{y} \\
\ddot{\varphi}
\end{array}\right)=\left(\begin{array}{ll}
\frac{c_{\alpha f}+C_{\alpha r}}{m v_{x}} & \frac{a C_{\alpha f} - b C_{\alpha r}}{m v_{x}}-V_{x} \\
\frac{a \cdot c_{\alpha f} - b C_{\alpha r}}{I \cdot V_x} & \frac{a^{2} C_{\alpha f}+b^{2} \cdot C_{\alpha r}}{I \cdot V_{x}}
\end{array}\right) \cdot\left(\begin{array}{l}
\dot{y} \\
\dot{\varphi}
\end{array}\right)+\left(\begin{array}{c}
-\frac{C_{\alpha f}}{m} \\
-\frac{a C_{\alpha r}}{I}
\end{array}\right) \cdot \delta
\end{equation}

Until now, the form of a nonlinear control system for two wheels bicycle model has already been provided, similar to $\dot{X} = A \cdot X + B \cdot u$. Here $u = \delta$. By adjusting the input of control $u$, we could control $y$ and $\varphi$, stabilizing the system. In the next part, Linear Quadratic Regular (LQR) will be used for the lateral control in the path following module.

\section{Control Algorithm}
\label{Control Algorithm}
Normally the control system of an autonomous vehicle could be divided into longitudinal control and lateral control. The component controls the longitudinal motion of the ego vehicle, considering its longitudinal dynamics. The controlled variables in this case are throttle and brake inputs to the ego vehicle, which govern its motion including velocity, acceleration, jerk and higher derivatives in the longitudinal direction. As for lateral control, the controlled variable, in this case, is the steering input to the ego vehicle, which governs its steering angle and heading. It is important to mention that the steering angle and heading are different. The heading is related to the orientation of the ego vehicle, while the steering angle describes the orientation of the steerable wheels and hence the direction of motion of the ego vehicle. 
\subsection{Lateral Control}
In the last section, we have already obtained the lateral control system of the 2-wheels bicycle model $\dot{X} = AX + Bu$,
and the error of lateral control is 
\begin{equation}
    \Vec{e}_{rr} = \Vec{x} - \Vec{x}_r
\end{equation}

$\Vec{x}_r$ is already given by the reference trajectory, and $\Vec{x}$ should satisfy the equation above. According to Control Theorem, it is easy to obtain:
\begin{equation}
    \dot{e}_{rr} = \Bar{A}e_{rr} + \Bar{B}u
\end{equation}

\begin{figure}[!t]\centering
	\includegraphics[width=13.5cm]{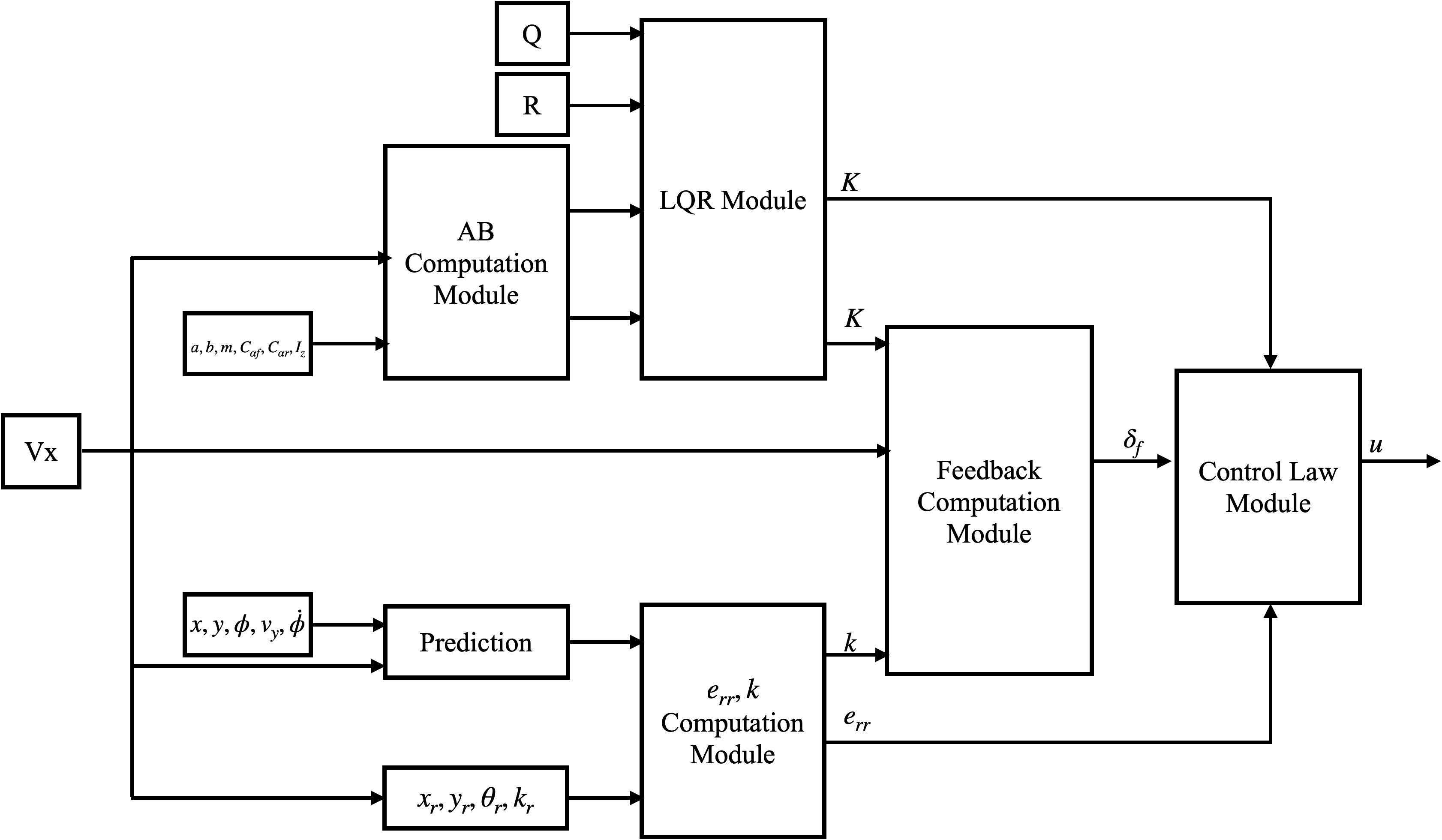}
	\caption{Lateral Control Structure}\label{Lateral Control Structure}
	\end{figure}
The objective of lateral control is to decide $u$ to make sure $\Vec{x}$ and $\Vec{x_r}$ close enough, which means to minimize $|\Vec{e_{rr}}|$.
Output
\begin{equation}
    u = -Ke_{rr} + \delta_f
\end{equation}

\begin{enumerate}
    \item Step 1: A, B matrices are computed by the computation module with $v_x$ and the parameters of vehicles. Then K will be computed with A, B, Q, and R as input.
    \item Step 2:According to the position and station, first do the prediction for the next step, then compute $e_{rr}$ and $k$.
    \begin{equation}
        \dot{e_{rr}} = A \cdot e_{rr} + B \cdot (-K \cdot e_{rr} + \delta_f) + C \cdot \dot{\theta}_r
    \end{equation}
    After stabilizing, $\dot{e}_{rr} = 0$, and 
    \begin{equation}
        e_{rr} = -(A - B \cdot K)^{-1} \cdot (B\cdot \delta_f + C \cdot \dot{\theta}_r)
    \end{equation}
    
    \item Step 3:Compute $\delta_f$
    Our objective is to choose $\delta_f$, so as to make $e_rr$ to be 0 as soon as possible.
    \begin{equation}
e_{r r}=\left(\begin{array}{c}
\frac{1}{K_{1}}\left\{\delta_{f}-\frac{\dot{\theta}_{r}}{v_{x}}\left[a+b-b \cdot K_{3}-\frac{m \cdot v_{x}^{2}}{a+b}\left(\frac{b}{c_{f}}+\frac{a}{c_{r}} \cdot K_{3}-\frac{a}{c_{r}}\right)\right]\right\} \\
0 \\
-\frac{\dot{\theta}_{r}}{v_{x}}\left(b+\frac{a}{a+b} \cdot \frac{m v_{x}^{2}}{c_{\alpha r}}\right) \\
0
\end{array}\right)
\end{equation}
So $e_d$ will become 0 when:
\begin{equation}
\delta_{f}=\frac{\dot{\theta}_r}{v_{x}}\left[a+b-b \cdot K_{3}-\frac{m \cdot v_{x}^{2}}{a+b}\left(\frac{b}{c_f}+\frac{a}{c_r} \cdot K_{3}-\frac{a}{c_r}\right)\right]
\end{equation}
$e_\varphi$ could not be influenced by $\delta_f$, so the system is not controllable. However, by estimation, we get the result that $e_\varphi = -\beta$, so $e_\varphi$ is almost a constant and equals to $-\beta$.  
    \item Step 4:\begin{equation}
        u = -Ke_{rr} + \delta_f
    \end{equation}

\end{enumerate}

We could define a cost function $J$ related to $e_{rr}$ and $u$
\begin{equation}
    J = e_{rr}^TQe_{rr} + u^TRu
\end{equation}

Q,R are the symmetric matrix, $e_{rr}$ and $u$ are the column vectors
The constraints for this cost function is
\begin{equation}
    \dot{e}_{rr} = \Bar{A}e_{rr} + \Bar{B}u
\end{equation}

The constraint is linear form, therefore it represents L in LQR. L plus QR is called the LQR algorithm in control theory.

Here the proof is neglected because it is easy to find it in any traditional Control theory book. The result of $e_{rr}$ is given here:
\begin{equation}
\begin{pmatrix}
\dot{e}_d \\ \Ddot{e}_d \\ \dot{e}_\phi \\ \Ddot{e}_\phi
\end{pmatrix}
= \begin{pmatrix}
0 & 1 & 0 & 0 \\
0 & \frac{C_{\alpha f} + C_{\alpha r}}{mv_x} & -\frac{C_{\alpha f} + C_{\alpha r}}{m} & \frac{a C_{\alpha f} - b C_{\alpha r}}{mv_x} \\
0 & 0 & 0 & 1 \\ 
0 & \frac{a C_{\alpha f} - b C_{\alpha r}}{Iv_x} & -\frac{a C_{\alpha f} - b C_{\alpha r}}{I} & \frac{a^2 C_{\alpha f} + b^2 C_{\alpha r}}{Iv_x}
\end{pmatrix}
\begin{pmatrix}
e_d \\ \dot{e}_d \\ e_\phi \\ \dot{e}_\phi
\end{pmatrix}
+ \begin{pmatrix}
0 \\ -\frac{C_\alpha f}{m} \\ 0 \\ -\frac{a C_{\alpha f}}{I}
\end{pmatrix} \delta
+ \begin{pmatrix}
0 \\ \frac{aC_{\alpha f} - b C_{\alpha r}}{mv_x} -v_x \\ 0 \\
\frac{a^2 C_{\alpha f} + b^2 C_{\alpha r}}{Iv_x}
\end{pmatrix} \dot{\theta}_r
\end{equation}

$\dot{\theta}_r$ could be neglected here because $\theta_r$ changes very small and $\dot{\theta}_r$ is almost 0.
So we have the form:
\begin{equation}
    \dot{e}_{rr} = \bar{A}e_{r} + \bar{B}u
\end{equation}
During this equation, $e_{rr}$ equals to $(\dot{e}_d, \Ddot{e}_d, \dot{e}_\phi, \Ddot{e}_\phi)^T$, $u$ equals to steering angle $\delta$. Until now, our problem has already been transformed into a linear control system, with matrix $\bar{A}$ and $\bar{B}$. By adjusting $u$, it is easy to make the system tend to be $0$, or minimize $e_rr$ in a limited time.
\subsection{Longitudinal Control}
As for longitudinal control, one of the easiest control laws - PID is chosen because of its simplicity and feasibility. PID is so easy and common that we will not discuss it in more detail about it here.

\section{Simulation and Results}
\label{Simulation and results}
\subsection{Simulation environment}
In a real implementation, Matlab/Simulink is the main tool for motion planning and controlling algorithm. Besides Simulink, CARSim is used to offer the dynamic module of vehicles. Prescan is chosen because it can offer a huge variety of sensors and simulation environments. The most important thing is both of the simulation software offer an interface with Matlab/Simulink. The Simulink module and the initialization file of the Motion Planning and Controlling part have already been uploaded to my GitHub: \href{https://github.com/Liyucheng1997/Motion-planning-with-adaptive-sampling.git}{link with all the models and codes in my experiment}. 
\subsection{Adaptive sampling visualization}
\begin{figure}[!t]\centering
	\includegraphics[width=13.5cm]{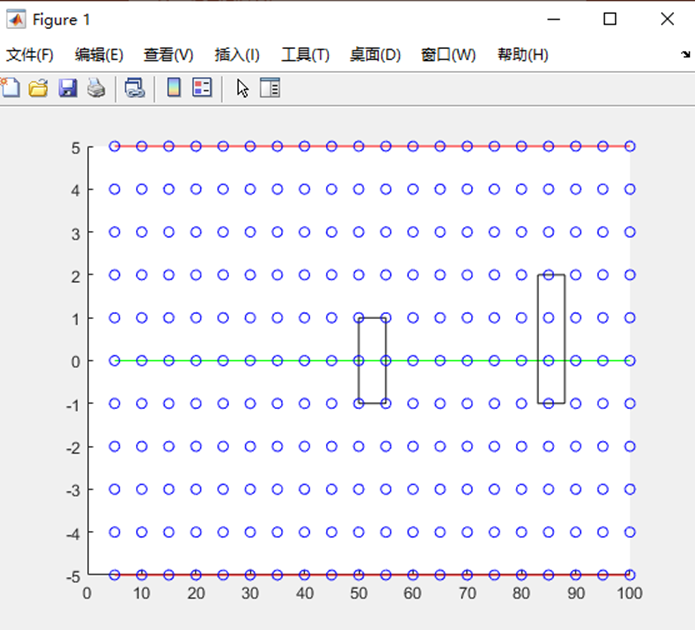}
	\caption{Average Sampling}\label{Average Sampling}
	\end{figure}

Here we used a scatter graph for the visualization of our Adaptive Sampling with an Artificial Potential Field. Firstly, an average sampling is implemented to divide the continuous space into discrete, as shown in Figure \ref{Average Sampling}

\begin{figure}[!t]\centering
	\includegraphics[width=13.5cm]{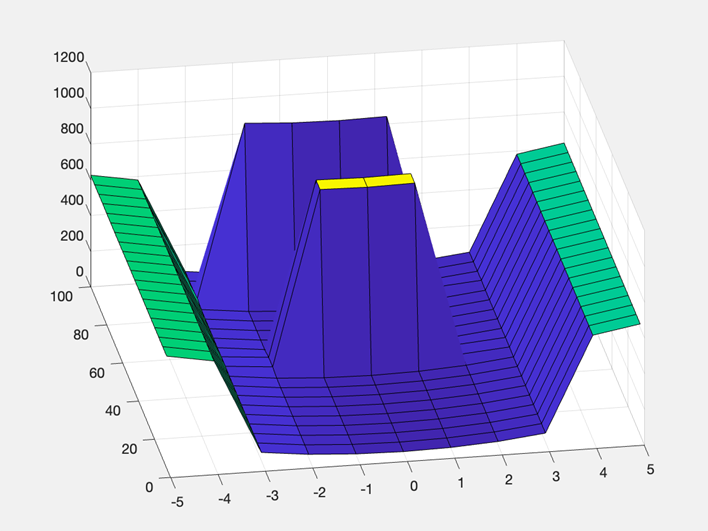}
	\caption{Adaptive Sampling Cost Visualization}\label{Adaptive Sampling Cost Visualization}
	\end{figure}

\begin{figure}[!t]\centering
	\includegraphics[width=13.5cm]{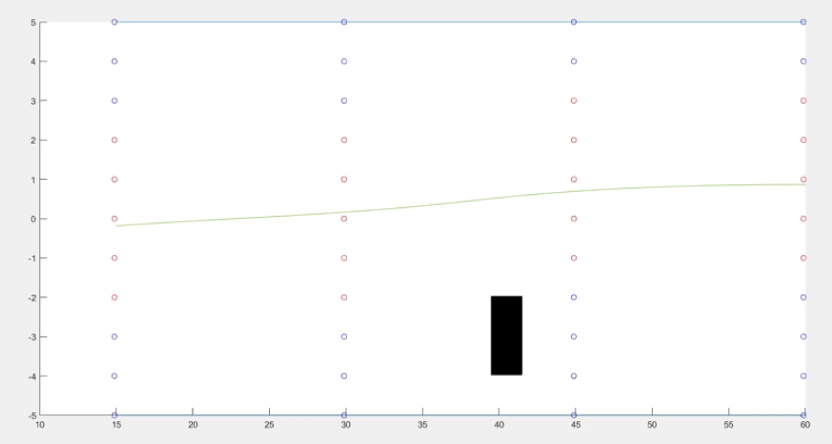}
	\caption{Adaptive Sampling}\label{Adaptive Sampling vis}
    \end{figure}

Then, in the next, ASAPF is used to calculate the cost for all the nodes in average sampling depending on their distance with an optimal trajectory in the last cycle, obstacles and road boundaries. One example of the cost is shown in Figure \ref{Adaptive Sampling Cost Visualization}. The label [-5,5] refers to the road width (10 meters). [0, 100] means the planning length is 100 meters. The value in the third dimension represents the cost of every single node in average sampling. Then, we just chose several nodes with a smaller cost in each station and build our adaptive sampling area. Our methods will delete the unnecessary points in average sampling, saving the running time of dynamic programming in motion planning without scarifying the performance of planning trajectory, as shown in Figure \ref{Adaptive Sampling vis}.

\subsection{Result comparison}
\subsubsection{SL graph computation}
\begin{figure}[!t]\centering
	\includegraphics[width=13.5cm]{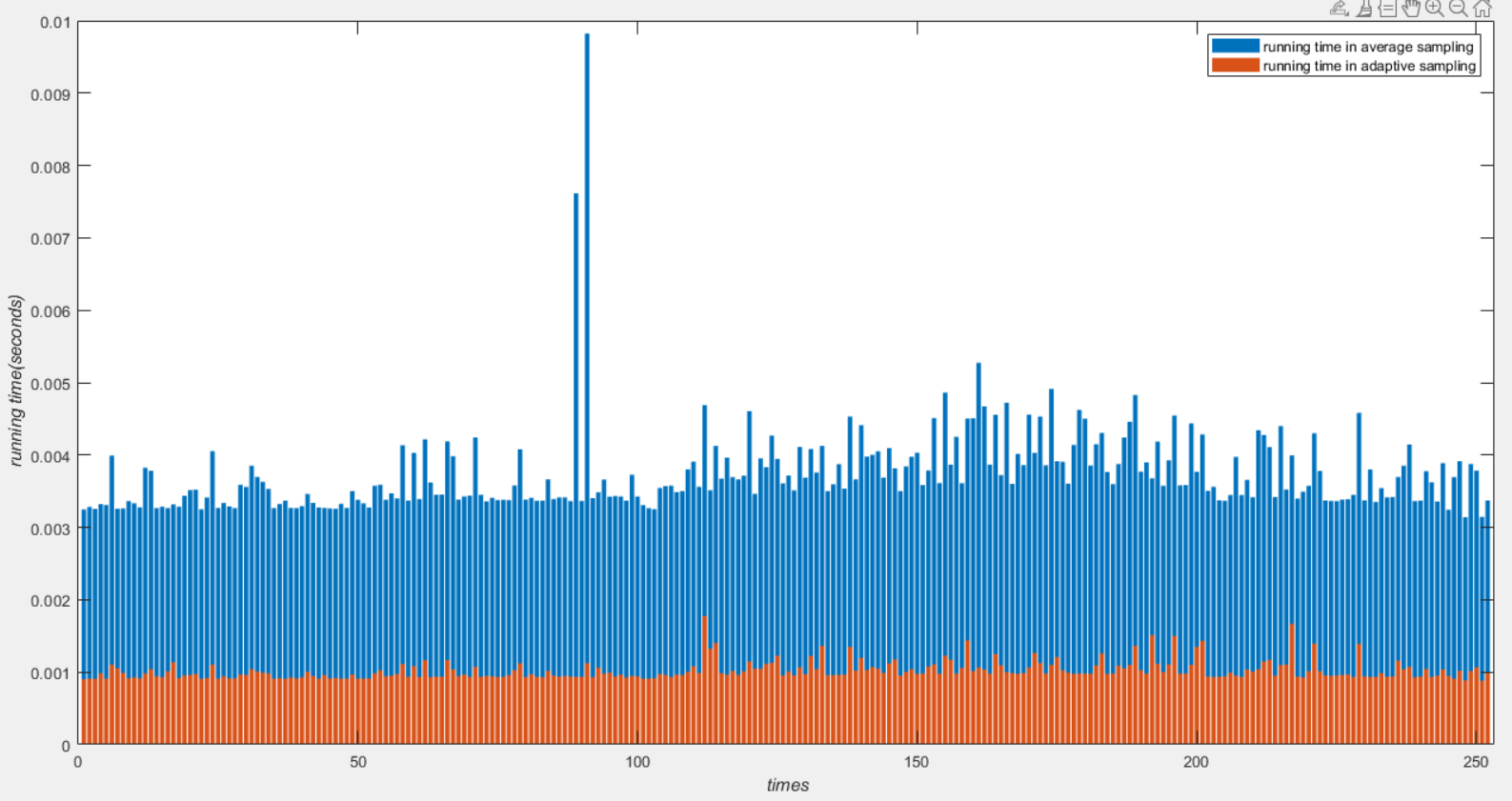}
	\caption{SL computation comparison}\label{SL computation comparison}
	\end{figure}

The experiment result is shown in Figure \ref{SL computation comparison}. The blue bar graph means the computation time of dynamic programming with original average sampling, and the orange bars represent the result of our ASAPF. The results clearly show that the running time of programming is hugely decreased, around $66.7\%$ average. In the next Figure \ref{Path performance comparison}, the optimal path performance in original average sampling and in adaptive sampling are compared together. The value we chose to compare is the minimal cost of dynamic programming which means the optimal path computed by programming. The result is almost the same which means the path performance is not influenced by our ASAPF. Therefore, it shows clearly that our Adaptive Sampling Method is feasible and useful to decrease the running time of dynamic programming without scarifying the performance of optimal solution at the same time.
\begin{figure}[!t]\centering
	\includegraphics[width=13.5cm]{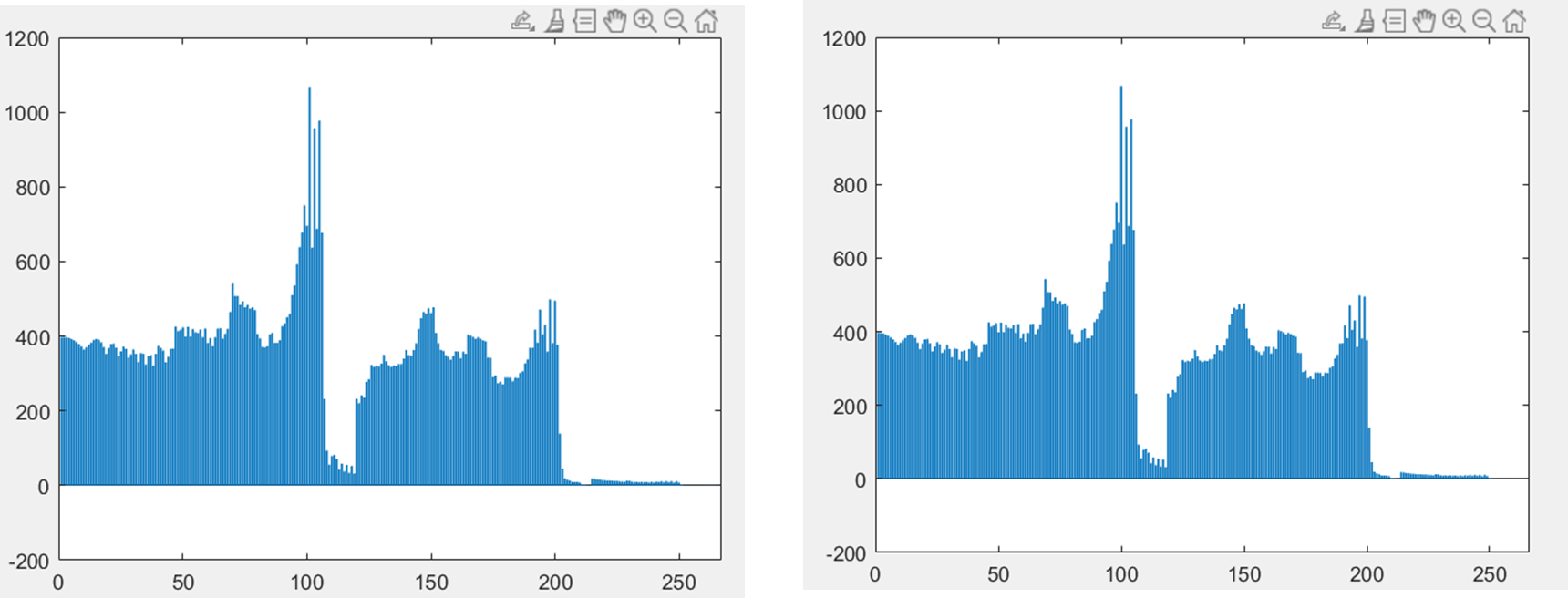}
	
	\caption{Path performance comparison}\label{Path performance comparison}
	\end{figure}
\subsubsection{ST graph computation}
\begin{figure}[!t]\centering
	\includegraphics[width=13.5cm]{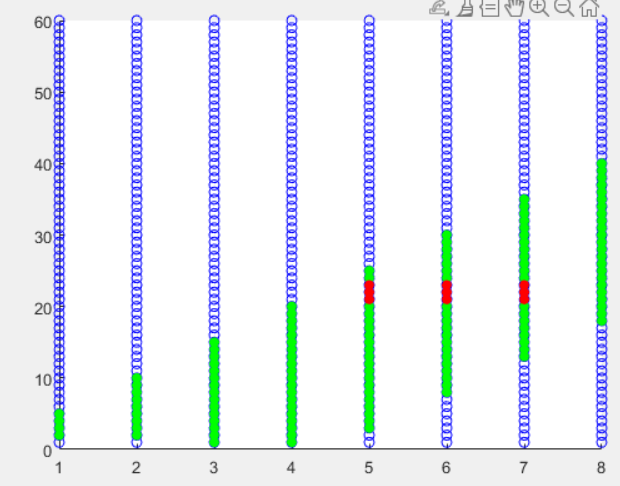}
	\caption{Adaptive Sampling in ST graph}\label{Adaptive Sampling performance in ST graph}
	\end{figure}

\begin{figure}[!t]\centering
	\includegraphics[width=13.5cm]{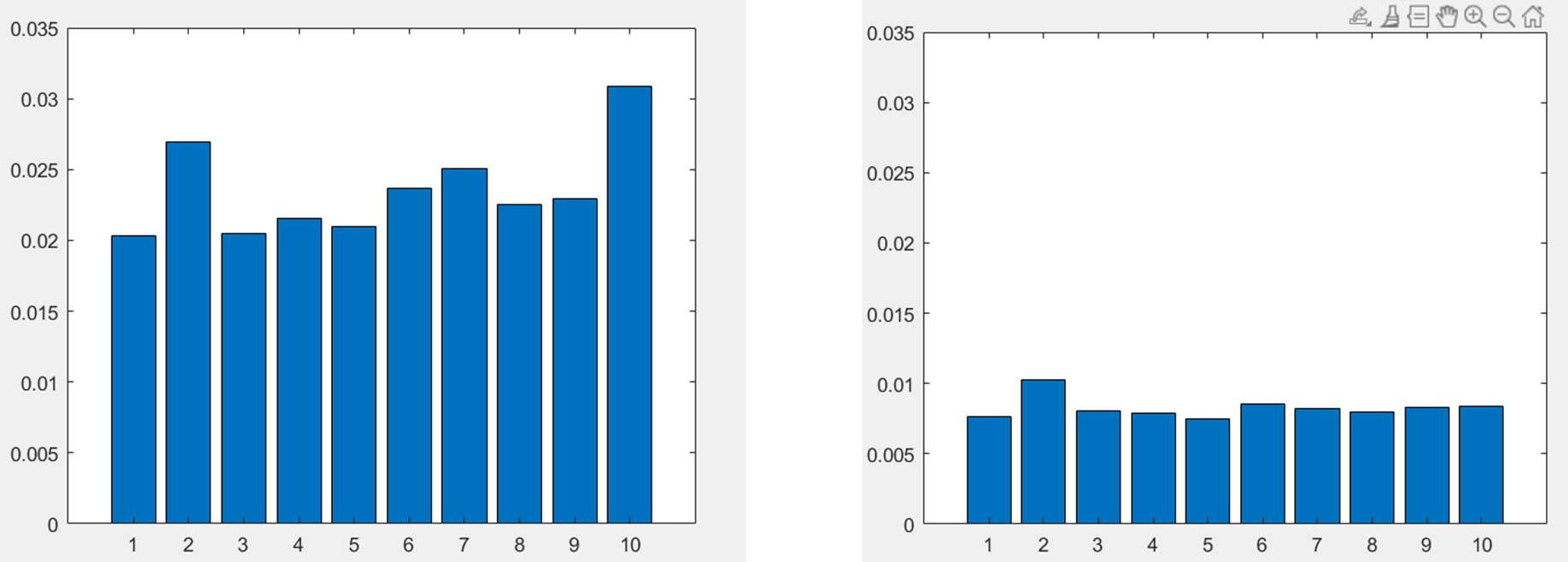}
	\caption{ST computation comparison}\label{ST computation comparison}
	\end{figure}

In the ST graph, Adaptive Sampling is also implemented, aiming to delete unnecessary points depending on the vehicle constraints. As shown in Figure \ref{Adaptive Sampling performance in ST graph}, with adaptive sampling, the green nodes mean the real feasible area for the next dynamic programming to solve the problem (red nodes mean obstacles). The ST computation time comparison is shown in Figure \ref{ST computation comparison}, the left bar graph represents the original average sampling, and the right graph means the running time of dynamic programming with our adaptive sampling in the ST graph. By comparison, it is easy to prove the feasibility of our method in saving running time and improving the performance of the motion planning algorithm in automated driving. 

\newpage
\section{Conclusions and Future Works}

\label{Conclusion and future works}
\subsection{Conclusion}
The objective of this thesis has been to speed up the dynamic programming for the motion planning of automated driving, using adaptive sampling to replace the conventional average sampling. An adaptive sampling has been proposed and formulated on the basis of the information on obstacles, road boundaries, prior optimal trajectory in the last cycle, and vehicle dynamic constraints. Based on the motion planning algorithm in Baidu Apollo - EM Planner, we have implemented the motion planning and control algorithm in MATLAB/Simulink. They have been co-simulated with Prescan and CARSim, which offer the model of sensors, dynamic vehicles and simulation environment, the model in Simulink proved the feasibility of our proposed adaptive sampling and the running time of dynamic programming has been reduced by 2/3 while keeping the same performance of final optimal trajectory. The improvement is significant for the motion planning module for the reason that the efficiency of the algorithm is very important for real-time implementation and the less time the algorithm needs to run, the more time the automated vehicle will have to react to some accident occasions and the safer the whole automated driving process will be. 

\subsection{Future work}

Due to the limitation of the time span of this  thesis, there are still some points that can be improved and are here indicated as future work.

To simplify the prediction of dynamic vehicles, the speed of obstacle vehicles was assumed constant in simulation tests, which is not the case in real cases. In this regard, a tool usually used relies on Neural Networks. For example, RNN or LSTM are commonly used to train a model to predict the trajectory of dynamic vehicles in many recent papers. However, more study is necessary to accurately predict the speed and trajectory of dynamic obstacles with such a tool. In addition, Game Theory is recently studied in automated driving problems, because other drivers in dynamic vehicles or other autonomous vehicles have their manoeuvres too when encountering other vehicles. The problem that they will yield or overtake should also be taken into consideration in the automated driving process, otherwise, there will be a failure of automated driving which will lead to heavy traffic accidents and be dangerous to human society. Last but not least, Dynamic programming has already been proven feasible for motion planning problems, but not the best solution for this problem. Another algorithm such as RRT*, or control theory (MPC) should also be studied in my future period to compare the performance of different methods. 
\newpage
\section*{Acknowledgements}

This is a master's thesis, finished at L'Aquila University, supported by Epico (Electric Pulsion and Control), an Erasmus Mundus + Program. This program includes four universities: Ecole Centrale de Nantes in France, L'Aquila University in Italy, Kiel Univerity in Germany, and the University in Romania. I spent the first semester in ECN, Nantes, France. Then, the last three semesters are spent in UAQ, L'Aquila, Italy. My tutor in UAQ is Professor Stefano Di Gennaro and my co-tutor is Professor Maarouf Saad from ETS, Canada. This master's thesis lasts around 4 months, from March to July. In the beginning, It was arranged to finish this thesis in Canada with Professor Maarouf Saad in present. However, due to the limitation of time and the visa issue, I could not go to Canada, so we changed our plan and then I decided to stay in L'Aquila to continue my thesis. Every Friday, we have a meeting with Professor Stefano and Professor Maarouf online in a Zoom meeting. Also, another Indian student whose name is Mohit is also supervised by these two professors. These two professors are really responsible for us and helped us a lot during this period. They give me a lot of suggestions and advice which helped me improve a lot in research. During this period, I found my interest in researching. They gave me enough freedom to find my interest in automated driving, and according to their knowledge, offered me sincere advice every time. I really appreciate the process of my master's thesis in the last four months. Therefore, I would like to continue my PhD study in the next. There are a lot of problems in the automated driving area I would like to research more and try my best to solve them. For example, as mentioned in future work, the prediction of other vehicles and pedestrians still remains a difficult topic to be solved in automated driving. My dream is to create a fully autonomous vehicle, L4 or even L5 level. It will be my pleasure if I could try my best to contribute a little bit in this area of automated driving. In my mind, technologies and science are like magic because they could do something others think impossible. In the last ten years, smartphones and the Internet are the hot topics. In the next ten years, automated driving and IoT will become the future. Based on my experience and knowledge, I am really happy that I could join in this evolution.

\setcounter{chapter}{7}
\chapter{Appendices}
\setcounter{section}{8}

\section{Appendix}
\subsection{Appendix 1}
\label{Appendix 1}
First of all, the definition of a jerk is given here. Jerk equals the rate of change of acceleration in time intervals related to the comfort of passengers in the real implementation of automated driving. In other words, if the trajectory is too jerky, the passengers will feel uncomfortable. Therefore, our main objective is to minimize the jerk:
$$J(t) = \dddot{f}(t)$$
If function $f(t)$ is quadratic or linear, definitely jerk will be minimal. However, there are six constraints the curve connecting two points needs to satisfy: 

    \begin{equation}
        \begin{cases}
        s(0) = s_0,  & s(T) = s_n; \\
        \dot{s}(0) = v_0, &  \dot{s}(T) = v_n; \\
        \ddot{s}(0) = a_0, & \ddot{s}(T) = a_n\\
    \end{cases}
    \end{equation}
    
Then, it is easy to obtain that:

\begin{equation}
    \begin{cases}
        s(T) - s(0) = s_n - s_0, \\
        \dot{s}(t) - \dot{s}(0) = v_n - v_0, \\
        \ddot{s}(T) - \ddot{s}(0) = a_n - a_0;
    \end{cases}
\end{equation}
Let us note that $s_n - s_0 = C_0$, $v_n - v_0 = C_1$ and $a_n - a_0 = C_2$. Then, the constraints will become:
\begin{equation}
    \begin{cases}
        s(T) - s(0) = \int \dot{f}dt =C_1,  \\
        \dot{s}(t) - \dot{s}(0) = \int \ddot{f}dt = C_2, \\
        \ddot{s}(T) - \ddot{s}(0) = \int \dddot{f}dt = C_3;
    \end{cases}
\end{equation}
    
Until now, the problem has already become how to solve problem $\int_{0}^{T} \dddot{f}(t)$ under the constraints 
\begin{equation}
    \begin{cases}
        \int_{0}^{T}(\dot{f} - \frac{C_{0}}{T})dt = 0 \\
        \int_{0}^{T}(\ddot{f} - \frac{C_{1}}{T})dt = 0 \\
        \int_{0}^{T}(\dddot{f} - \frac{C_{2}}{T})dt = 0
    \end{cases}
\end{equation}

Here Euler-Lagrange Equation is used to solve this problem:

\begin{equation}
    \int_{0}^{T} \dddot{f}^2dt + \lambda_1*\int_{0}^{T}(\dot{f} - \frac{C_0}{T})dt + 
\lambda_2*\int_{0}^{T}(\ddot{f} - \frac{C_1}{T})dt +
\lambda_3*\int_{0}^{T}(\dddot{f} - \frac{C_2}{T})dt 
= \int_{0}^{T} (\lambda_1\dot{f} + \lambda_2\dddot{f} + \dddot{f}^2 - \lambda_1\frac{C_0}{T} - \lambda_2\frac{C1}{T} - \lambda_3\frac{C2}{T})dt 
= \int_{0}^{T} L dt (L = L(\dot{f}, \ddot{f},\dddot{f}))
\end{equation}

Euler-Lagrange Equation tells us the objective function will have minimal value if the equation below could be satisfied:
\begin{equation}
    \frac{\partial L}{\partial f} - \frac{\mathrm{d}}{\mathrm{d}t}\frac{\partial L}{\partial \dot{f}} + \frac{\mathrm{d}^2}{\mathrm{d}t^2}\frac{\partial L}{\partial \ddot{f}} - \frac{\mathrm{d}^3}{\mathrm{d}t^3}\frac{\partial L}{\partial \dddot{f}} = 0
\end{equation}

By the expression of $L$, we could have:
$\frac{\partial L}{\partial f} = 0$, 
$\frac{\partial L}{\partial \dot{f}} = \lambda_1$,
$\frac{\partial L}{\partial \dddot{f}} = \lambda_2$,
$\frac{\partial L}{\partial \dddot{f}} = \lambda_3 + 2\dddot{f}$
Then, the final result is obtained like that:
$\frac{\mathrm{d}^3}{\mathrm{d}t^3}(\lambda_3 + 2\dddot{f}) = 0$
Let us go back to deduce the expression of $f$ from this result, hence,
   \begin{equation}
       \begin{cases}
        f^{(6)}(t) = 0, \\
        f^{(5)}(t) = a_{0}, \\
        f^{(4)}(t) = a_{0}t+a_{1}, \\
        f^{(3)}(t) = \frac{1}{2}a_{0}t^2 + a_1t + a_{2}, \\
        f^{(2)}(t) = \frac{1}{6}a_{0}t^3 + \frac{1}{2}a_{1} t^2 + a_2t + a_{3}, \\
        f^{(1)}(t) = \frac{1}{24}a_0t^4 + \frac{1}{6}a_1t^3 + \frac{1}{2}a_2t^2 + a_3t + a_{4}, \\
        f^{(0)}(t) = \frac{1}{120}a_0t^{5} + \frac{1}{24}a_1t^{4} + \frac{1}{6}a_2t^{3} + \frac{1}{2}a_3t^{2} + a_{4}t + a_{5}
        \end{cases}
   \end{equation}
    
Until now, we have already proved why a fifth-degree polynomial could satisfy the requirement of a minimal jerk and it is also the minimum degree that can satisfy our standard in motion planning problem.

\subsection{Appendix 2: Frenet-Cartesian Coordinate Transformation}
\label{Appendix 2}

First of all, the definitions of $\dot{s}$, $\dot{l}$ and $l^{'}$ are given here:
\begin{equation}
    \dot{s} = \frac{ds}{dt},
    \dot{l} = \frac{dl}{dt}, 
    l^{'} = \frac{dl}{ds}
\end{equation}
It is important to mention that $\dot{l}$ and $l^{'}$ could be transformed to each other easily, following the equations below:
\begin{equation}
    \label{transfer_equation}
    \dot{l} = \frac{dl}{dt} = \frac{dl}{ds}\frac{ds}{dt} = l^{'}\dot{s}
\end{equation}
Then, seven basic equations are needed before the real deduction.
\begin{equation}
    \label{basic equation 1}
    \Vec{\dot{r_n}} = \Vec{v} = |\Vec{v}|\Vec{\tau_n}
\end{equation}

\begin{equation}
    \label{basic equation 2}
    \Vec{\dot{r_r}} = \dot{s}\Vec{\tau_r}
\end{equation}

\begin{equation}
\label{basic equation 3}
    \Vec{\dot{\tau_n}} = k_n|\Vec{v}|\Vec{n_n}
\end{equation}

\begin{equation}
\label{basic equation 4}
    \Vec{\dot{n_n}} = -k_n|\Vec{v}|\Vec{\tau_n}
\end{equation}

\begin{equation}
\label{basic equation 5}
    \Vec{\dot{\tau_r}} = k_r\dot{s}\Vec{n_r}
\end{equation}

\begin{equation}
\label{basic equation 6}
    \Vec{\dot{n_r}} = -k_r\dot{s}\Vec{n_r}
\end{equation}

\begin{equation}
\label{basic equation 7}
    \Vec{a} = |\Vec{\dot{v}}|\Vec{\tau_n} + |\Vec{v}|^2k_n\Vec{n_n}
\end{equation}

The core equation is this:
\begin{equation}
    \label{core equation}
    \Vec{r_r} + l\Vec{n_r} = \Vec{r_n}
\end{equation}

\begin{enumerate}
    \item $s$ \\
    The nearest reference point in the reference road line can be expressed as $\vec{r} = [x_r, y_r] $. The $s_r$ in this reference point will be chosen as $s$ of $[x_x, y_x]$ in Frenet frame, therefore we have:
    \begin{equation}
    s = s_r
    \end{equation}
    \item $\dot{s}$ \\
    Do derivative to both sides in Equation \ref{core equation}, we could obtain:
    \begin{equation}
         \Vec{\dot{r_r}} + l\Vec{\dot{n_r}} + \dot{l}\Vec{n_r}= \Vec{\dot{r_n}}
    \end{equation}
    Combined with \ref{basic equation 2}, \ref{basic equation 6} and \ref{basic equation 3}, then the equation becomes:
    \begin{equation}
        \dot{s}\Vec{\tau_r} + l(-k_r\dot{s}\Vec{\tau_r}) + \dot{l}\Vec{n_r} = \Vec{v}
    \end{equation}
    
    Then, Make dot product with $\vec{\tau_r}$ in both sides, it becomes: $$\dot{s} + l(-k\dot{s}) = \dot{v}\vec{\tau_r} $$
    So, we achieved the final expression of $\dot{s}$:
    \begin{equation}
    \label{dot_s}
            \dot{s} = \frac{\dot{v}\vec{\tau_r}}{1-k_rl}
    \end{equation}
    
    \item $l$ \\
    Transform core Equation \ref{core equation}, it is not difficult to obtain:
    \begin{equation}
        l\Vec{n_r} = \Vec{r_n} - \Vec{r_r} 
    \end{equation}
    
    So,
    \begin{equation}
        l = (\Vec{r_n} - \Vec{r_r})\Vec{n_r}
    \end{equation}

    \item $\dot{l}$ \\
    \begin{equation}
    \label{l_dot}
       \dot{l} = \vec{v}\vec{n_r} 
    \end{equation}
    \item $l^{'}$ \\
    \begin{equation}
    \label{l_prime}
       l^{'} = \frac{dl}{ds} = \frac{\dot{l}}{\dot{s}} = \frac{\vec{v}\vec{n_r}}{\frac{\vec{v}\vec{\tau_r}}{1-k_rl}} = (1-k_rl)\frac{\vec{v}\vec{n_r}}{\vec{v}\vec{\tau_r}}   
    \end{equation}
  
    \item $\ddot{s}$ \\
    Based on the expression of $\dot{s}$ in Equation \ref{dot_s}, we could do the derivative to get $\ddot{s}$:
    \begin{equation}
\begin{aligned}
&\ddot{s}=\frac{d \dot{s}}{d t}=\frac{1}{\left(1-k_{r} l\right)^{2}}\left(\frac{d\left(\vec{v} \cdot \overrightarrow{\tau_{r}}\right)}{d t} \cdot\left(1-k_{r} l\right)-\left(\vec{v} \cdot \overrightarrow{\tau_{r}}\right) \cdot\left(-\dot{k}_{r}l - -k_{r} \cdot \dot{l}\right)\right)\\
&=\frac{1}{1-k_{r l}}\left(\frac{d \vec{v}}{d t} \cdot \overrightarrow{\tau_{r}}+\vec{v} \cdot \frac{d \overrightarrow{\tau_{r}}}{d t}\right)+\frac{1}{1-k_{r} l} \cdot \frac{\overrightarrow{v} \cdot \overrightarrow{\tau_{r}}}{1-k_{r} l}\left(\dot{k_{r}} \cdot l+k_{r} \cdot \dot{l}\right) .\\
&=\frac{1}{1-k_{r} \cdot l} \lvert(\vec{a} \cdot \overrightarrow{\tau_{r}}+\vec{v} \cdot \lvert(k_{r} \cdot \dot{s} \cdot \overrightarrow{n_{r}} \lvert)+\frac{1}{1-k_{r} \cdot l} \cdot \dot{s} \cdot \lvert(\frac{d k_{r}}{d s} \cdot \frac{d s}{d t} \cdot l \cdot k_{r} \cdot \frac{d l}{d s} \cdot \frac{d s}{d t} \lvert) \lvert \\
&=\frac{\vec{a} \cdot \overrightarrow{\tau_{r}}}{1-k_{r} l}+\frac{\left(k_{r} \cdot \dot{s}\right) \cdot\left(\vec{v} \cdot \overrightarrow{n_{r}}\right)}{1-k_{r} \cdot l}+\frac{\dot{s}^{2}}{1-k_{r} l}\left(k_{r}^{\prime} l \cdot+k_{r} \cdot l^{\prime}\right)
\end{aligned}
\end{equation}

Combined with Equation \ref{l_dot} and \ref{transfer_equation}, the expression of $\ddot{s}$ will be obtained:
\begin{equation}
\ddot{s}=\frac{\vec{a} \cdot \overrightarrow{\vec{\tau}_r}}{1-k_{r} \cdot l}+\frac{k_{r} \cdot \dot{s}^{2} \cdot l^{\prime}}{1-k_{r} l}+\frac{\dot{s}^{2}}{1-k_{r} l}\left(k_{r}^{\prime} \cdot l+k_{r} \cdot l^{\prime}\right)
\end{equation}

    \item $\ddot{l}$ \\
    From Equation \ref{l_dot}, we could have the relation between $\dot{l}$ and $\vec{v}$, then make the derivative of both sides in the equation, we will have:
    \begin{equation}
\begin{aligned}
\ddot{l} &=\frac{d \vec{v}}{d t} \cdot \overrightarrow{n_{r}}+\vec{v} \cdot \frac{d \overrightarrow{n_{r}}}{d t} \\
&=\vec{a} \cdot \overrightarrow{n_{r}}+\vec{v} \cdot\left(-k_{r} \cdot \dot{s} \cdot \overrightarrow{\tau_{r}}\right) \\
&=\vec{a} \cdot \overrightarrow{n_{r}}-k_{r} \cdot \dot{s}\left(\vec{v} \cdot \overrightarrow{\tau_{r}}\right)
\end{aligned}
\end{equation}
Because of Equation \ref{basic equation 5}, we have the relation 
\begin{equation}
    \Vec{\dot{\tau_r}} = k_r\dot{s}\Vec{n_r}
\end{equation}
Therefore, 
\begin{equation}
    \ddot{l}=\vec{a} \cdot \overrightarrow{n_{r}}-k_{r}^{2} \cdot \dot{s}^{2} \cdot \dot{l} 
\end{equation}
    \item $l^{''}$ \\
    \begin{equation}
\begin{aligned}
\ddot{l}=\frac{d \dot{l}}{d t}=\frac{d\left(l^{\prime} \cdot \dot{s}\right)}{d t} &=\frac{d l^{\prime}}{d t} \cdot \dot{s}+l^{\prime} \cdot \ddot{s} \\
&=\frac{d l^{\prime}}{d s} \cdot \frac{d s}{d t} \cdot \dot{s}+l^{\prime} \cdot \ddot{s} \\
&=l^{\prime \prime} \cdot \dot{s}^{2}+l^{\prime} \cdot \ddot{s}
\end{aligned}
\end{equation}
so,
\begin{equation}
l^{\prime \prime}=\frac{\ddot{l}-l^{\prime} \cdot \ddot{s}}{\dot{s}^{2}}
\end{equation}
\end{enumerate}

\backmatter
\cleardoublepage
\phantomsection

\bibliographystyle{plain}
\bibliography{refs}
\title{Abstract}
\end{document}